\documentclass[3p,review]{elsarticle}

\usepackage{lineno,hyperref}
\usepackage{natbib}
\usepackage{adjustbox}
\usepackage{longtable}
\usepackage{booktabs}
\usepackage{xcolor}
\usepackage{caption}
\usepackage{rotating}
\usepackage{lscape}
\usepackage{soul}
\usepackage{textcomp}
\usepackage{multirow}
\usepackage{multicol}
\usepackage{tabu}
\usepackage{tocloft}
\usepackage{blindtext}
\usepackage{mathtools}
\usepackage{blkarray}
\usepackage{subcaption}
\usepackage{float}
\usepackage{array}
\usepackage{amsmath}
\usepackage{amssymb}
\usepackage{setspace}
\usepackage{bm}
\usepackage{algorithm}
\usepackage{algpseudocode}
\usepackage{makecell}

\begin{document}

\begin{frontmatter}

\title{Sustainable Greenhouse Microclimate Modeling: A Comparative Analysis of Recurrent and Graph Neural Networks}
% \tnotetext[mytitlenote]{Fully documented templates are available in the elsarticle package on \href{http://www.ctan.org/tex-archive/macros/latex/contrib/elsarticle}{CTAN}.}


\author[1]{\corref{cor1} Emiliano Seri}
\ead{emiliano.seri@uniroma2.it}
\cortext[cor1]{Corresponding author}

\author[2]{Marcello Petitta}
\ead{marcello.petitta@uniroma3.it}

\author[3]{Chryssoula Papaioannou}
\ead{chpapa@uth.gr}

\author[4]{Nikolaos Katsoulas}
\ead{nkatsoul@uth.gr}

\author[1]{Cristina Cornaro}
\ead{cristina.cornaro@uniroma2.it}

\address[1]{Department of Enterprise Engineering, University of Rome Tor Vergata, Via del Politecnico 1, 00133, Rome, Italy}

\address[2]{Department of Mathematics and Physics, Roma Tre University, Via della Vasca Navale 84, 00146, Rome, Italy}

\address[3]{Department of Agrotechnology, University of Thessaly, 41500 Larissa, Greece}

\address[4]{Lab of Agricultural Constructions and Environmental Control, Department of Agriculture Crop Production and Rural Environment, University of Thessaly, Fytokou Str, 38446, Volos, Greece}

\begin{abstract}

The integration of photovoltaic (PV) systems into greenhouses not only optimizes land use but also enhances sustainable agricultural practices by enabling dual benefits of food production and renewable energy generation.
However, accurate prediction of internal environmental conditions is crucial to ensure optimal crop growth while maximizing energy production. 
This study introduces a novel application of Spatio-Temporal Graph Neural Networks (STGNNs) to greenhouse microclimate modeling, comparing their performance with traditional Recurrent Neural Networks (RNNs). 
While RNNs excel at temporal pattern recognition, they cannot explicitly model the directional relationships between environmental variables. Our STGNN approach addresses this limitation by representing these relationships as directed graphs, enabling the model to capture both environmental dependencies and their directionality. 
We benchmark RNNs against directed STGNNs on two 15-min-resolution datasets from Volos (Greece): a six-variable 2020 installation and a more complex eight-variable greenhouse monitored in autumn 2024.  In the simpler 2020 case the RNN attains near-perfect accuracy, outperforming the STGNN.  
When additional drivers are available in 2024, the STGNN overtakes the RNN ($R^{2}=0.905$ vs $0.740$), demonstrating that explicitly modelling directional dependencies becomes critical as interaction complexity grows.  
These findings indicate when graph-based models are warranted and provide a stepping-stone toward digital twins that jointly optimise crop yield and PV power in agrivoltaic greenhouses.

\end{abstract}

% Highlights section

\begin{highlights}
\item Compare RNNs and directed STGNNs for greenhouse microclimate prediction.
\item RNN excels in the simple 2020 scenario.
\item STGNN beats RNN on a complex 2024 greenhouse with eight environmental drivers.
\item Directional graphs help when added variables amplify interaction paths.
\item Result guides when to pick temporal vs. spatio-temporal models.
\end{highlights}

\begin{keyword}
Recurrent neural networks \sep Spatio-temporal graph neural networks \sep Graph attention networks \sep Long short-term memory networks \sep Greenhouse modeling.
\end{keyword}

\end{frontmatter}

% \linenumbers

\section{Introduction}
\label{sec:intro}

Modern agriculture faces increasing pressure to optimize land use while reducing its environmental impact. The integration of photovoltaic (PV) systems into agricultural greenhouses has emerged as a promising solution to these challenges, offering dual benefits of food production and renewable energy generation. However, this integration introduces new complexities in greenhouse management, as PV panels directly influence light distribution, temperature patterns, and overall microclimate dynamics. These Agri-PV systems optimize land use by combining solar energy harvesting with protected crop cultivation, offering potential benefits for both energy and food security. Yet the successful implementation of PV-integrated greenhouses requires accurate prediction and control of the internal microclimate to ensure optimal growing conditions while maximizing energy production. While sophisticated physical models incorporating greenhouse thermodynamics, crop physiology, and PV system performance exist \citep{ZAINALI2025125558}, their complexity often restricts real-time applications and rapid design optimization. This study, conducted within the framework of the European REGACE project, represents an initial step toward developing a comprehensive digital twin for PV-integrated greenhouses.
Rather than relying on complex physical models, we explore empirical approaches that employ advanced machine learning to predict greenhouse microclimates.
By comparing traditional Recurrent Neural Networks (RNNs) with directed Spatio-Temporal Graph Neural Networks (STGNNs), we investigate whether explicit modeling of environmental relationships and directional influences can improve prediction accuracy while maintaining computational efficiency.

RNNs capture temporal patterns but treat each variable independently, whereas STGNNs embed each environmental variable as a node in a directed graph, allowing the model to learn how changes propagate through explicit causal pathways \citep{Yu2018,Velickovic2018,SUN2022119739}. 
We analyse 15-min data from two Mediterranean greenhouses in Volos (Greece): GH2 (2020, six variables, $\approx$33 k samples across three seasons) and the more complex GH4 (autumn 2024, eight variables including PAR and CO\textsubscript{2}, 4 k training / 1 k test). Both models use identical hyperparameter budgets and a 96-step (24 h) input window, ensuring that performance differences arise from the modelling paradigm rather than tuning effort. A vanilla RNN achieves near-perfect accuracy on GH2, yet the STGNN overtakes it on GH4 ($R^{2}=0.905$ vs 0.740), demonstrating that directional graphs earn their keep when additional sensors amplify interaction pathways. This finding addresses the open question of when graph machinery is necessary for greenhouse prediction, complementing earlier empirical studies that relied solely on temporal models \citep{SEGINER1994203,FERREIRA200251,DEmilio2012NeuralNF,FOURATI20071016,Gharghory2020,HONGKANG2018790,Manonmani2018ModellingAC,gao2023temperature}. 
It also advances digital-twin research for PV-integrated houses by showing how physics-informed edge sets can improve data-driven predictions without exhaustive search. Effective micro-climate control remains challenging because internal temperature and humidity respond non-linearly to ambient weather, solar radiation and ventilation events \citep{AASLYNG2005521,AASLYNG2003657}, and PV panels introduce further heterogeneity \citep{OUAZZANICHAHIDI2021116156,BAGLIVO2020115698,BRAEKKEN2023101830}. Hybrid, physics-based simulations capture those processes \citep{STANCIU2016498,ABDELGHANY20061521,MOBTAKER201988,SINGH2018227,FITZRODRIGUEZ2010105,MASSA2011711} but are often too slow for on-line use; our results show that a graph-augmented neural network can bridge the gap when variable interactions are rich.

The paper is structured as follows: Section~\ref{sec: Data} describes the data used in this study, including data collection methods, preprocessing steps, and the handling of missing values. Section~\ref{sec: method} outlines the methodologies employed, offering a concise introduction to deep learning and detailing the architectures and implementation of the RNN and directed STGNN models. Section~\ref{sec: Analysis} presents the analysis and results, comparing the performance of the RNN and directed STGNN models in predicting greenhouse temperature across different seasons. Finally, Sections~\ref{sec: Discussion} and \ref{sec: conclusions} discuss the findings, their implications for greenhouse management, and potential directions for future research.

\section{Data}
\label{sec: Data}

This study draws on two high-frequency (15-min) data sets collected in Volos (Greece) under the REGACE project.  
\textbf{GH2} spans an entire year (January to December 2020; 33 446 records) with six environmental variables, whereas \textbf{GH4} is an autumn 2024 campaign (9 October to 20 November 2024; 5 000 records) that adds PAR and CO\textsubscript{2} sensors.

\subsection*{GH2 (2020) – six-variable benchmark}  
The features considered during the analysis are given below. The labels used during the analysis are shown in parentheses:
\begin{itemize}
    \item External Environmental Factors:
    \begin{itemize}
        \item External Temperature (OUT\_temp): The ambient temperature outside the greenhouse.
        \item External Relative Humidity (OUT\_RH): The relative humidity of the air outside the greenhouse.
        \item Radiation Level (OUT\_rad): The amount of solar radiation received.
        \item Wind Speed (OUT\_wind\_speed): The speed of the wind outside the greenhouse.
    \end{itemize}
    \item Internal Environmental Factors:
    \begin{itemize}
        \item Greenhouse Temperature (G2\_temp): The internal temperature within Greenhouse.
        \item Greenhouse Relative Humidity (G2\_RH): The internal relative humidity within Greenhouse.
    \end{itemize}
\end{itemize}

\begin{figure}[H]
    \centering
    \caption{GH2 temperatures inside and outside the greenhouse for the autumn period}
\label{fig: autumntemp}
\includegraphics[width=1\textwidth]{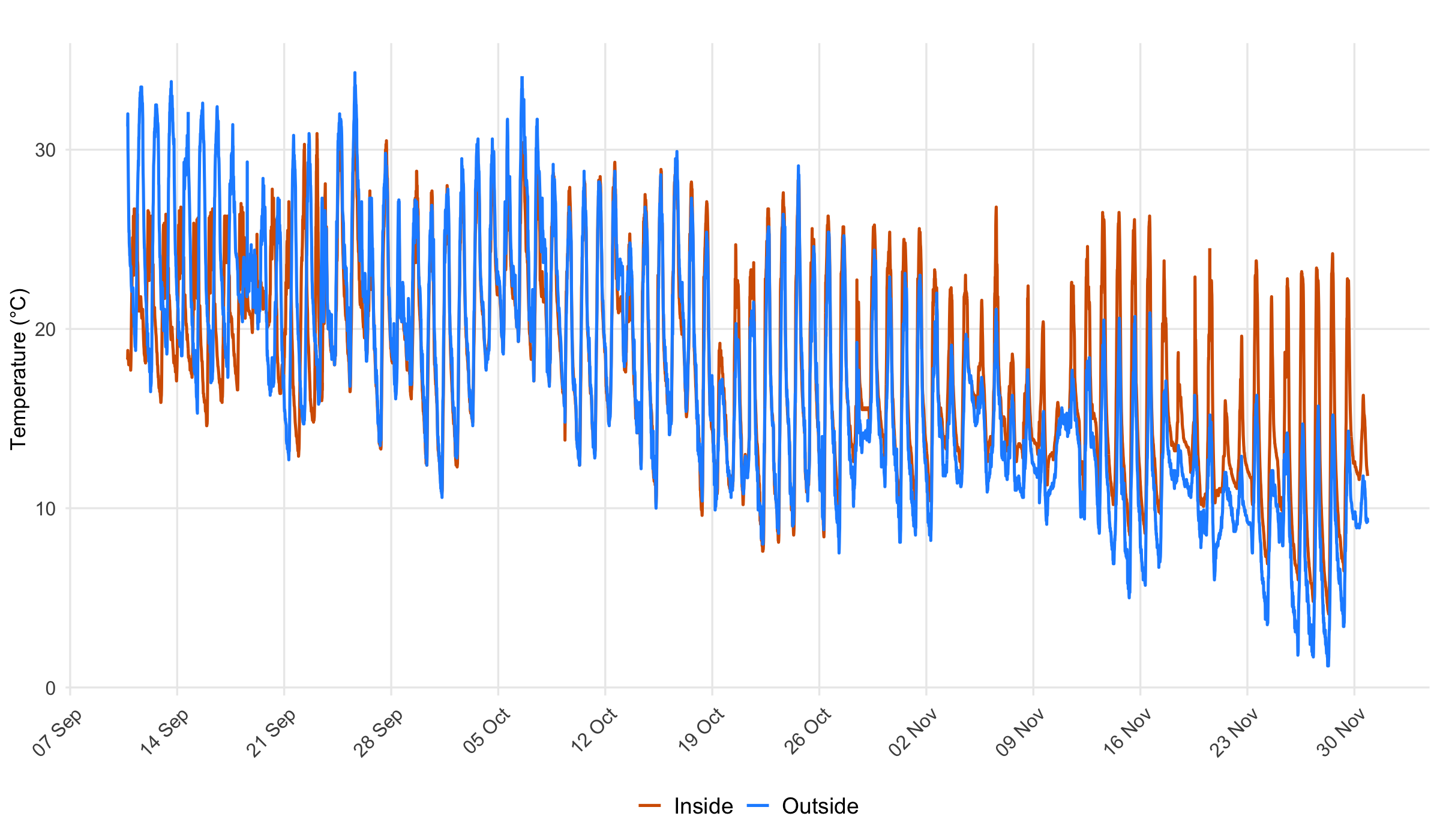}
\end{figure}

The inside-outside temperature trajectories for the autumn period in GH2 is shown in Figure \ref{fig: autumntemp}. The winter and summer periods for GH2 are shown in \ref{fig: wintertemp} and \ref{fig: summertemp} in Appendix \ref{secAppendixSeasonal}. During the summer period, the greenhouse employs a cooling system to prevent internal temperatures from rising excessively. Spring was excluded because of the high number of missing values.  For modelling each season is split chronologically: the first 80 \% of records form the training set, the remaining 20 \% the test set.

\begin{figure}[H]
\centering
\caption{GH2 feature-interaction graph}
\label{fig:Graph}
\includegraphics[width=\textwidth]{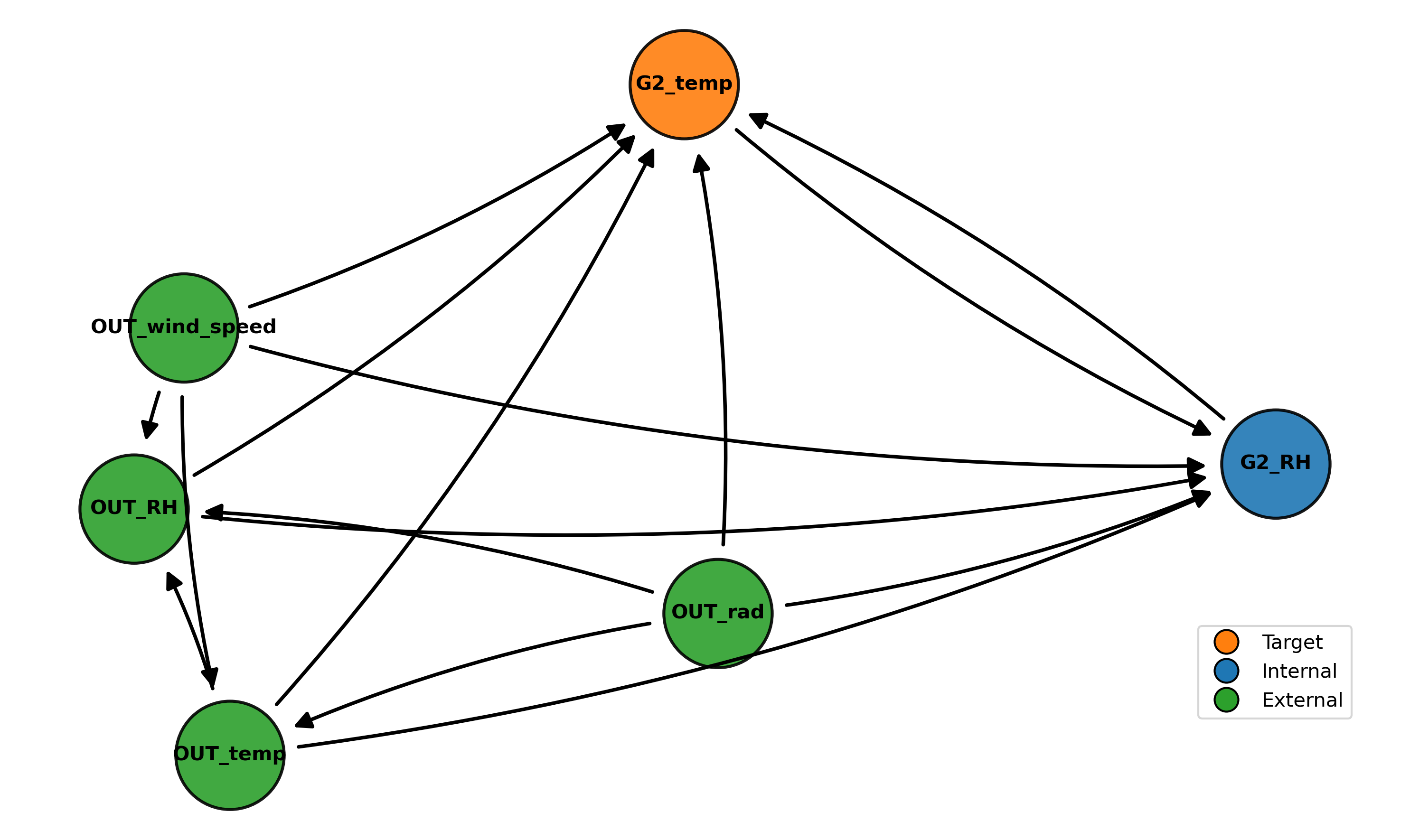}
\end{figure}

The directed edges in our graph structure represent fundamental physical relationships between environmental variables, based on well-established thermodynamic and atmospheric physics principles. Our choice of edge directions is motivated by the following physical mechanisms:

\begin{enumerate}
    \item External Relative Humidity Dependencies:
    \begin{itemize}
        \item Temperature $\rightarrow$ RH: Controls air's water vapor capacity through the Clausius-Clapeyron relation
        \item Wind Speed $\rightarrow$ RH: Affects vapor mixing and transport through turbulent diffusion
        \item Solar Radiation $\rightarrow$ RH: Influences evaporation rates and local temperature gradients
    \end{itemize}
    \item External Temperature Dependencies:
    \begin{itemize}
        \item Solar Radiation $\rightarrow$ Temperature: Primary driver through radiative heating
        \item Wind Speed $\rightarrow$ Temperature: Modifies heat exchange through forced convection
    \end{itemize}
    \item Internal Parameters (Temperature and RH) Dependencies:
    \begin{itemize}
        \item External Temperature $\rightarrow$ Internal Temperature: Heat transfer through greenhouse envelope
        \item External RH $\rightarrow$ Internal RH: Vapor exchange through ventilation and infiltration
        \item Solar Radiation → Internal Parameters: Direct heating and greenhouse effect
        \item Wind Speed $\rightarrow$ Internal Parameters: Convective heat exchange rate
        \item Internal Temperature $\leftrightarrow$ Internal RH: Bidirectional coupling through:
        \begin{itemize}
            \item Temperature affecting air's water-holding capacity
            \item Humidity influencing evaporative cooling effects
            \item Plant transpiration processes
        \end{itemize}
    \end{itemize}
\end{enumerate}
These physical relationships form the basis for the graph structure shown in Figure~\ref{fig:Graph}, where each edge represents a direct influence between variables. This directed graph approach allows our STGNN model to learn and weight these physical dependencies during the prediction process.

\subsection*{GH4 (2024) – eight-variable greenhouse}
Compared with GH2 the 2024 installation is instrumented with two additional physical dimensions: photosynthetically active radiation (PAR) and carbon-dioxide concentration, both measured inside and outside the structure:

\begin{itemize}
\item (G4\_PAR) / (OUT\_PAR): PAR is the spectral band plants use for photosynthesis.
\item (G4\_CO2) / (OUT\_CO2): dry‐air CO\textsubscript{2} mixing ratio (ppm). 
\end{itemize}

\begin{figure}[H]
    \centering
    \caption{GH4 temperatures inside and outside the greenhouse for the autumn period}
\label{fig: GH4autumntemp}
\includegraphics[width=1\textwidth]{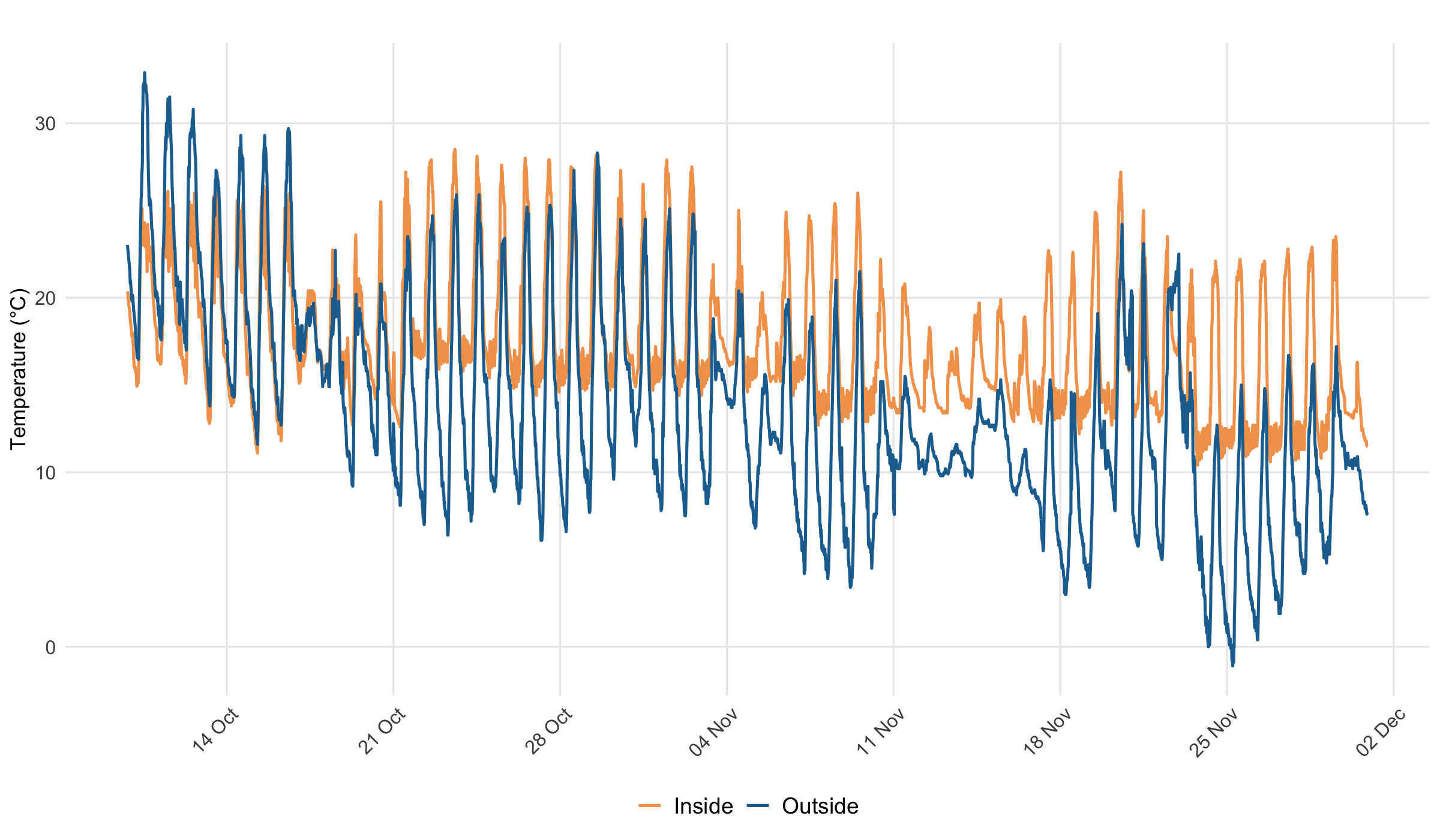}
\end{figure}

Together with the common variables (air temperature, relative humidity and broadband solar radiation, these sensors yield five external drivers (OUT\_Temp, OUT\_Rad, OUT\_PAR, OUT\_CO2, OUT\_RH) and three internal drivers (G4\_PAR, G4\_CO2, G4\_Temp [target]). Data cover a single autumn window and are split chronologically 80/20 for modelling (train: 9 Oct 20:00–20 Nov 11:45; test: 20 Nov 12:00–30 Nov 21:45). 
Figure~\ref{fig: GH4autumntemp} shows the inside-outside temperature trajectories for the autumn period in GH4.

\begin{figure}[H]
\centering
\caption{GH4 feature-interaction graph}
\label{fig:GraphGH4}
\includegraphics[width=\textwidth]{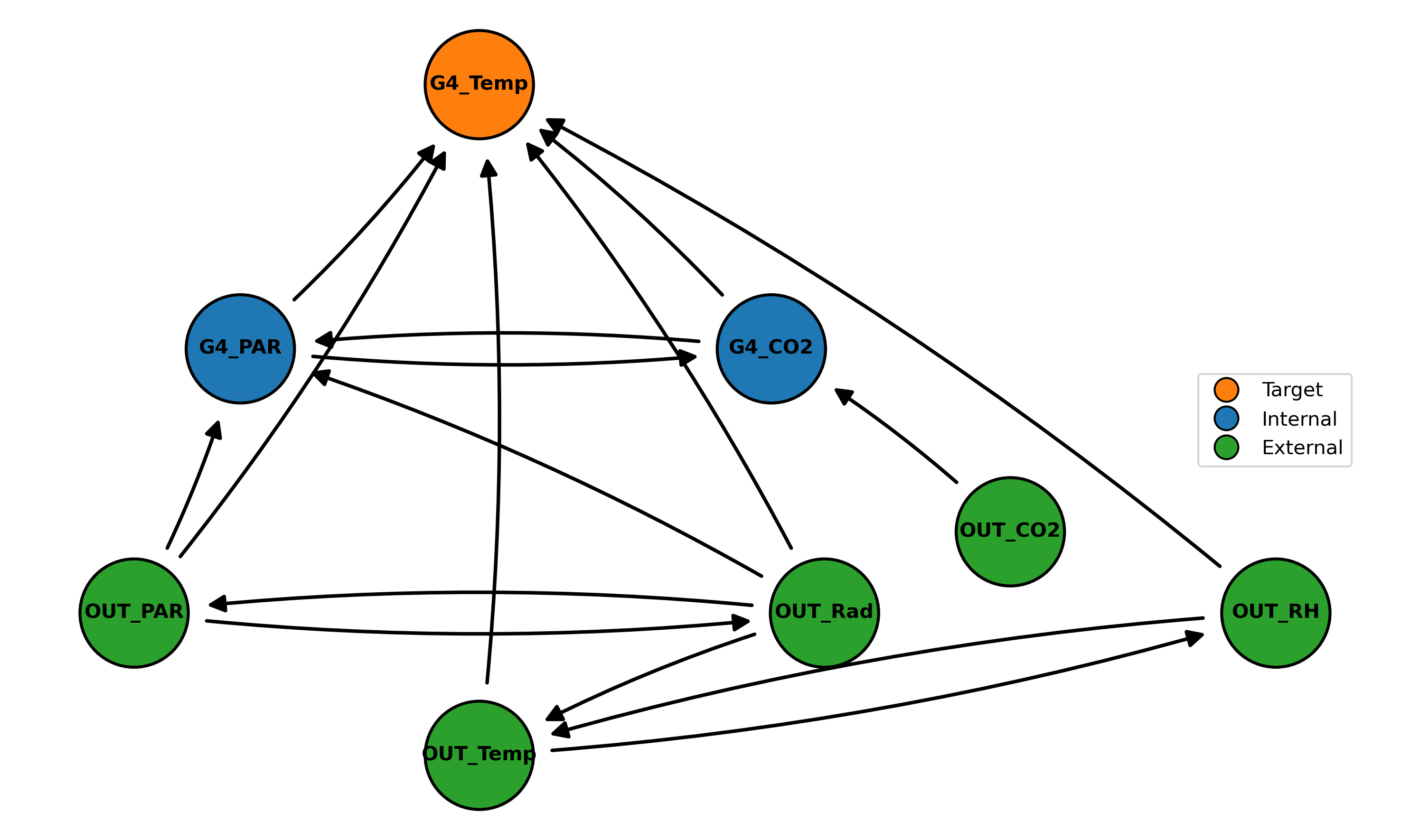}
\end{figure}

The rich PAR–CO\textsubscript{2} couplings motivate the denser edge set of Figure~\ref{fig:GraphGH4}, where directional links capture how radiation drives photosynthesis, which in turn modulates both internal CO\textsubscript{2} and latent-heat–induced humidity changes.

\subsection*{Missing-value treatment}  
GH2 contains 0–75 gaps per variable (winter: up to 68; autumn: 75). 
GH4 exhibits 273 missing points for every external variable and for G4\_PAR / G4\_CO2, and 274 for G4\_Temp and G4\_RH.  

Gaps are imputed with a \emph{directional five-point mean}: for a missing stamp at time $t$ (e.g.\ 13 Sep 01:00) we average the same quarter-hour on the two previous days ($t-24$ h, $t-48$ h) and the two following days ($t+24$ h, $t+48$ h).  
This approach ensured that the imputed values preserved the continuity in the time series without introducing noticeable distortions or flattening in the data sequences.

\subsection*{Normalisation and stationarity}  
All features are scaled to $[0,1]$ using min–max normalisation \citep{alaimo2021monitoring}.  Augmented Dickey–Fuller tests confirm stationarity for every variable (all $p<0.01$), so no further differencing is applied \citep{Cheung1995, kirchgassner2012introduction, tsay2005analysis}.

\section{Methodology}
\label{sec: method}

Neural Networks (NNs) are computational models inspired by the human brain's neural structure. They consist of interconnected nodes (neurons) organized into layers: an input layer, one or more hidden layers, and an output layer. Each connection between neurons has an associated weight that is adjusted during training to minimize prediction error. Deep Neural Networks (DNNs) extend NNs by incorporating multiple hidden layers, enabling the modeling of complex, nonlinear relationships in data. Activation functions such as ReLU, sigmoid, and tanh introduce nonlinearity, while backpropagation adjusts the weights based on the error rate obtained in previous epochs \citep{nielsen2015neural}.

This section outlines the deep learning techniques employed to develop predictive models of greenhouse temperature. We focus on two architectures: Recurrent Neural Networks (RNNs) and directed Spatio-Temporal Graph Neural Networks (STGNNs). 
We discuss their theoretical foundations, how they are applied in this study, and why they are particularly well-suited for modeling greenhouse microclimates.

\subsection{Recurrent neural networks}
\label{sub: RNNmethod}

Recurrent Neural Networks (RNNs) are specialized neural networks designed to process sequential data, making them highly suitable for time-series analysis \citep{JAIN1999}. Unlike traditional feedforward neural networks, RNNs have recurrent connections that allow information to persist across time steps, enabling the network to capture temporal dependencies.

An RNN processes sequences one element at a time, maintaining a hidden state that captures information about previous inputs. 
The hidden state $\bm{h}_t$ at time step $t$ is updated based on the current input $\bm{x}_t$ and the previous hidden state $\bm{h}_{t-1}$:
\begin{equation} 
\bm{h}_t = \tanh(\bm{W}_{h} \bm{x}_{t} + \bm{U}_{h} \bm{h}_{t-1} + \bm{b}_{h}) \end{equation}
The output $\hat{y}_t$ at time step $t$ is produced from the hidden state $\bm{h}_t$:
\begin{equation} 
\hat{y}_t = \bm{W}_y \bm{h}_t + \bm{b}_y \end{equation}
where $\bm{W}_h$, $\bm{U}_h$, and $\bm{W}_y$ are weight matrices. $\bm{b}_h$ and $\bm{b}_y$ are bias vectors.
$\tanh$ is the hyperbolic tangent activation function introducing nonlinearity.
In deeper RNNs, multiple RNN layers are stacked, passing the hidden state from one layer to the next, allowing the model to capture more complex temporal patterns \citep{nketiah2023recurrent}.

RNNs are effective for modeling how current and past environmental conditions influence future states, making them ideal for predicting variables like greenhouse temperature \citep{RODRIGUEZ1999}. 
A known challenge with RNNs is the vanishing gradient problem, where gradients diminish during backpropagation through time, making it difficult to learn long-term dependencies \citep{BEHRANG20101468}. However, since our data exhibits short-term dependencies, this issue do not impact our study \citep{DUBININ2024106179}.

\subsubsection{RNN implementation}

We begin by preparing the dataset, where each input sequence $\bm{X}^{(i)}$ consists of historical environmental measurements over a period of $T$ time steps. The target value $y^{(i)}$ is the greenhouse temperature at the next time step.

The notation used in our RNN methodology is defined as follows:
\begin{itemize}
    \item \textbf{Datasets and Sequences:}
    \begin{itemize}
        \item $\bm{X}^{(i)} \in \mathbb{R}^{T \times F}$: Input sequence for the $i$-th sample, where $T$ is the sequence length and $F$ is the number of features.
        \item $y^{(i)} \in \mathbb{R}$: Target value for the $i$-th sample.
        \item $N$: Total number of training samples.
    \end{itemize}
    \item \textbf{Model Parameters:}
    \begin{itemize}
        \item $E$: Number of training epochs.
        \item $B$: Batch size.
        \item $\bm{\theta}$: Set of all learnable parameters in the RNN.
    \end{itemize}
    \item \textbf{Layers and Functions:}
    \begin{itemize}
        \item $\text{SimpleRNN}(u, \text{return\_seq})$: Simple RNN layer with $u$ units and an option to return sequences.
        \item $\text{Dropout}(r)$: Dropout layer with a dropout rate $r$.
        \item $\text{Dense}(o)$: Fully connected layer with $o$ output units.
        \item $\text{Loss}(\cdot, \cdot)$: Loss function (Mean Squared Error).
        \item $\text{Optimizer}$: Optimization algorithm (Adam optimizer).
    \end{itemize}
\end{itemize}

The RNN model is constructed with two Simple RNN layers, each containing 50 units. The first RNN layer is configured to return sequences $\texttt{(return\_sequences=True)}$, allowing the subsequent layer to receive the entire sequence output. Dropout layers with a rate of 0.2 are interleaved between the RNN layers to prevent overfitting by randomly setting a fraction of input units to zero during training. A Dense layer with one unit at the end of the network outputs the predicted temperature.

The model is compiled using the Adam optimizer \citep{kingma2014adam}, which adapts the learning rate during training for efficient convergence. 
The loss function is set to Mean Squared Error (MSE), suitable for regression tasks.

Training proceeds for $E$ epochs, where in each epoch, the model is trained on the entire training dataset using batches of size $B$. Optionally, a portion of the training data is set aside for validation to monitor the model's performance on unseen data during training.

After training, the model is evaluated on the test dataset to assess its predictive performance. Evaluation metrics such as MSE, Root Mean Squared Error (RMSE), and the coefficient of determination ($R^2$) are computed to quantify the accuracy of the model's predictions.

The training algorithm is summarized as follows:

\begin{algorithm}[H]
\caption{Recurrent Neural Network (RNN) Training Algorithm}
\begin{algorithmic}[1]
\Require Training sequences $\{(X^{(i)}, y^{(i)})\}_{i=1}^N$, sequence length $T$, batch size $B$, number of epochs $E$
\State \textbf{Define} model architecture:
\State \hskip1em \textbf{Layer 1}: Simple RNN with 50 units, input shape $(T, F)$, return sequences \textbf{True}
\State \hskip1em \textbf{Layer 2}: Dropout layer with rate $0.2$
\State \hskip1em \textbf{Layer 3}: Simple RNN with 50 units
\State \hskip1em \textbf{Layer 4}: Dropout layer with rate $0.2$
\State \hskip1em \textbf{Layer 5}: Dense layer with 1 unit (output)
\State \textbf{Compile} model with optimizer \texttt{Adam} and loss \texttt{Mean Squared Error}
\For{epoch $= 1$ to $E$}
    \State \textbf{Train} the model on training data with batch size $B$
    \State \textbf{Optionally validate} the model on validation data
\EndFor
\State \textbf{Evaluate} the model on test data
\State \textbf{Compute} evaluation metrics (MSE, RMSE, $R^2$)
\end{algorithmic}
\end{algorithm}

\subsection{Spatio-Temporal Graph Neural Networks}
\label{sub: GNNmethod}

To model the spatio-temporal dynamics of the greenhouse microclimate, we employ directed STGNN. This approach integrates Graph Attention Networks (GATs) for spatial feature extraction \citep{Velickovic2018} and Long Short-Term Memory (LSTM) networks \citep{Hochreiter1997} for temporal sequence modeling.
Unlike traditional RNN-based models, which focus primarily on temporal sequences, STGNNs leverage Graph Neural Networks (GNNs) to incorporate topological and relational information among variables. This is well-aligned with recent advances in GNNs for time series \citep{Jin2024}. By representing each environmental variable, such as internal temperature, humidity, and external factors like solar radiation and wind speed, as a node in a directed graph, we capture how changes in one variable may propagate to others through directional edges.
Figure~\ref{fig:Graph} and \ref{fig:GraphGH4} illustrates this architecture. Each node corresponds to an environmental variable, while edges encode the direction of influence. At each time step $t$, node features $\bm{X}_t$ are extracted from the input sequence $\bm{X}^{(i)}$. To prevent data leakage, the feature of the target node (greenhouse temperature) at the current time step is masked (set to zero) before model input. The GAT layers learn to weigh the importance of edges and neighboring nodes, revealing spatial dependencies, while the LSTM layers capture long-term temporal patterns.
More specifically, we adopt the GAT because its edge-wise attention coefficients are computed for each directed edge, allowing the model to weight asymmetric causal links (e.g., \textit{OUT\_PAR $\rightarrow$ G4\_PAR}) that undirected GCN or GraphSAGE aggregators would blur.
For the temporal branch we choose an LSTM cell: it offers the same gating mechanism as the Simple-RNN baseline (ensuring a fair comparison), remains numerically stable over our 96-step sequences, and is a well-established benchmark for reproducibility.
It is important to emphasise that the graph is topological in feature space rather than in physical space: each node represents an environmental variable (e.g., PAR, CO\textsubscript{2}, humidity), not a sensor location, and edges encode functional or causal dependencies among variables rather than geographical proximity.

\subsubsection{Graph Attention Networks and Directed Graphs}

The Graph Attention Network (GAT) layer \citep{Velickovic2018} is employed to handle the directed graph and assign different weights to incoming edges through an attention mechanism. This mechanism allows each node to weigh the importance of its neighbors' features when updating its own representation.

For each node $i$ and its incoming neighbor $j$, the attention coefficient $\alpha_{ij}$ is computed to determine the importance of node $j$'s features to node $i$:

\begin{equation}
e_{ij} = \text{LeakyReLU}\left( \bm{a}^\top \left[ \bm{W} \bm{x}_i \, \Vert \, \bm{W} \bm{x}_j \right] \right)
\end{equation}

\begin{equation}
\alpha_{ij} = \frac{\exp(e_{ij})}{\sum_{k \in \mathcal{N}_i} \exp(e_{ik})}
\end{equation}

where:

\begin{itemize}
    \item $\bm{x}_i$ and $\bm{x}_j$ are the feature vectors of nodes $i$ and $j$, respectively.
    \item $\bm{W}$ is a learnable weight matrix.
    \item $\bm{a}$ is a learnable weight vector (attention mechanism).
    \item $[ \cdot \, \Vert \, \cdot ]$ denotes the concatenation of two vectors.
    \item $\text{LeakyReLU}$ is the activation function introducing nonlinearity.
    \item $\mathcal{N}_i$ is the set of neighboring nodes sending messages to node $i$.
    \item $e_{ij}$ is the unnormalized attention score.
    \item $\alpha_{ij}$ is the normalized attention coefficient, representing the weight assigned to node $j$'s features by node $i$.
\end{itemize}

The LeakyReLU function, characterized by a small positive slope in its negative part, prevents zero gradients and ensures a more robust learning process, especially important in the attention mechanism where gradient flow is crucial for dynamic weight adjustments.

\textbf{Node Feature Update:}

The updated feature vector $\bm{h}_i$ for node $i$ is computed by aggregating the transformed features of its neighbors, weighted by the attention coefficients:

\begin{equation}
\bm{h}_i = \sigma\left( \sum_{j \in \mathcal{N}_i} \alpha_{ij} \bm{W} \bm{x}_j \right)
\end{equation}

where $\sigma$ is an activation function, such as ReLU.

By using the GAT layer, which inherently supports directed graphs, we effectively model the directionality of interactions among environmental variables. The attention mechanism allows the model to assign different weights to incoming edges based on their importance, capturing the dependence relationships in the data.

\subsubsection{Long Short-Term Memory (LSTM) Networks}

Long Short-Term Memory (LSTM) networks \citep{Hochreiter1997} are a specialized type of RNN architecture designed to better capture and retain long-term dependencies in sequential data. Unlike traditional RNN units, which can suffer from the vanishing gradient problem, LSTM cells incorporate gating mechanisms, namely the input, forget, and output gates, allowing the network to selectively add or remove information from its internal cell state. This structure enables the LSTM to preserve context over many time steps, making it well-suited for complex temporal sequences where relevant patterns may span extended intervals. In the directed STGNN model, the LSTM layer processes the sequence of GAT outputs $\{\bm{H}t\}_{t=1}^T$ to extract temporal features that complement the spatial dependencies learned by the GAT layer.

\subsubsection{Directed STGNN model implementation}

The notation used in our STGNN methodology is defined as follows:

\begin{itemize}
    \item \textbf{Datasets and Sequences:}
    \begin{itemize}
        \item $\bm{X}^{(i)} \in \mathbb{R}^{T \times N_n \times F}$: Input sequence for the $i$-th sample, where $T$ is the sequence length, $N_n$ is the number of nodes, and $F$ is the feature size per node (here, $F=1$).
        \item $y^{(i)} \in \mathbb{R}$: Target value for the $i$-th sample.
        \item $N$: Total number of training samples.
    \end{itemize}
    \item \textbf{Graph Components:}
    \begin{itemize}
        \item $E$: Edge index representing directed connections between nodes in the graph.
        \item $N_n$: Number of nodes in the graph.
    \end{itemize}
    \item \textbf{Model Parameters:}
    \begin{itemize}
        \item $E_p$: Number of training epochs.
        \item $B$: Batch size.
        \item $H$: Hidden size of the GAT layer.
        \item $H'$: Hidden size of the LSTM layer.
        \item $K$: Number of attention heads in the GAT layer.
        \item $\bm{\theta}$: Set of all learnable parameters in the STGNN.
    \end{itemize}
    \item \textbf{Layers and Functions:}
    \begin{itemize}
        \item $\text{GAT}(F, H, K)$: GAT layer with input feature size $F$, output size $H$ per head, and $K$ attention heads.
        \item $\text{LSTM}(H \times K \times N_n, H')$: LSTM layer with input size $H \times K \times N_n$ and hidden size $H'$.
        \item $\text{Dropout}(r)$: Dropout layer with a dropout rate $r$.
        \item $\text{Dense}(o)$: Fully connected layer with $o$ output units.
        \item $\text{Loss}(\cdot, \cdot)$: Loss function (Mean Squared Error).
        \item $\text{Optimizer}$: Optimization algorithm (Adam optimizer).
    \end{itemize}
\end{itemize}

Layer 1 (\textbf{GAT Layer}) processes node features $\bm{X}_t$ at each time step $t$ using the graph structure defined by $E$. It aggregates information from neighboring nodes based on attention coefficients to capture spatial dependencies.
Layer 2 (\textbf{LSTM Layer}) processes the sequence of GAT outputs $\{ \bm{H}_t \}_{t=1}^T$ to capture temporal dependencies over the sequence length $T$.
Layer 3 (\textbf{Dropout Layer}) is applied with a rate of 0.2 to mitigate overfitting.
Layer 4 (\textbf{Dense Output Layer}) produces the predicted greenhouse temperature $\hat{y}$.

To avoid data leakage, the feature of the target node (greenhouse temperature) at the current time step is set to zero during input preparation.
The model is compiled using the Adam optimizer \citep{kingma2014adam} and the Mean Squared Error (MSE) loss function. It is trained over $E_p$ epochs using batches of size $B$. During training, gradients are computed through backpropagation, and model parameters $\bm{\theta}$ are updated accordingly.

After training, the model is evaluated on the test dataset. Evaluation metrics such as MSE, RMSE, and $R^2$ score are calculated to assess the model's performance.

The training algorithm is summarized as follows:
\begin{algorithm}[H]
\caption{Directed STGNN Training Algorithm}
\begin{algorithmic}[1]
\Require Training sequences $\{ (\bm{X}^{(i)}, y^{(i)}) \}_{i=1}^N$, edge index $E$, sequence length $T$, batch size $B$, number of epochs $E_p$, number of nodes $N_n$
\State \textbf{Initialize} model parameters $\bm{\theta}$
\State \textbf{Define} model architecture:
\State \hskip1em \textbf{Layer 1}: $\text{GAT}(F, H, K)$
\State \hskip1em \textbf{Layer 2}: $\text{LSTM}(H \times K \times N_n, H')$
\State \hskip1em \textbf{Layer 3}: $\text{Dropout}(0.2)$
\State \hskip1em \textbf{Layer 4}: $\text{Dense}(1)$ (output layer)
\For{epoch $= 1$ to $E_p$}
    \For{each batch $\{ (\bm{X}_{\text{batch}}, y_{\text{batch}}) \}$}
        \State \textbf{Forward pass}:
        \For{time step $t = 1$ to $T$}
            \State Extract node features $\bm{X}_t$ from $\bm{X}_{\text{batch}}$ at time $t$
            \State \textbf{Prevent data leakage}: Set target node feature $\bm{x}_v \gets 0$
            \State Apply GAT layer: $\bm{H}_t \gets \text{GAT}(\bm{X}_t, E)$
        \EndFor
        \State Stack $\bm{H}_t$ over time to form $\bm{H}_{\text{seq}}$
        \State Apply LSTM layer: $\bm{H}_{\text{lstm}} \gets \text{LSTM}(\bm{H}_{\text{seq}})$
        \State Apply Dropout: $\bm{H}_{\text{drop}} \gets \text{Dropout}(\bm{H}_{\text{lstm}})$
        \State Compute output: $\hat{y} \gets \text{Dense}(\bm{H}_{\text{drop}})$
        \State \textbf{Compute loss}: $L \gets \text{Loss}(y_{\text{batch}}, \hat{y})$
        \State \textbf{Backward pass}: Compute gradients $\nabla_{\bm{\theta}} L$
        \State \textbf{Update parameters}: $\bm{\theta} \gets \bm{\theta} - \eta \nabla_{\bm{\theta}} L$
    \EndFor
    \State \textbf{Optionally validate} the model on validation data
\EndFor
\State \textbf{Evaluate} the model on test data
\State \textbf{Compute} evaluation metrics (MSE, RMSE, $R^2$)
\end{algorithmic}
\end{algorithm}

\section{Analysis and results}
\label{sec: Analysis}

Both architectures were trained and tested with a sequence length of $T=96$ time steps (a full 24-h cycle at 15-min resolution).  They share identical hyper-parameters: 32\,epochs, batch size~96, Adam optimiser, 10\,\% validation split, to ensure that any performance difference stems from the modelling paradigm rather than tuning effort \citep{GNmodelling2021,JUNG2020}.  For the STGNN we use a \emph{four-head} GAT layer ($K=4$), concatenating the heads so that each node outputs $H \times K$ features before the LSTM.

To compare the simple six-variable greenhouse (GH2) with the richer eight-variable PV house (GH4) we centre the discussion on the Autumn periods.  Summer and Winter results for GH2 are presented in Appendix~\ref{secAppendixSeasonal}.

\subsection*{GH2 (2020) versus GH4 (2024)}

\begin{table}[H]
\centering
\caption{Test performance in Autumn ($T\!=\!96$, 32\,epochs, batch 96)}
\label{tab:autumn_metrics}
\begin{tabular}{llccc}
\toprule
Greenhouse & Model & MSE & RMSE & $\bm{R^2}$ \\
\midrule
\multirow{2}{*}{GH2 (2020)} & RNN   & 0.0015 & 0.0393 & 0.958 \\
                            & STGNN & 0.0042 & 0.0650 & 0.884 \\[2pt]
\multirow{2}{*}{GH4 (2024)} & RNN   & 0.0110 & 0.1048 & 0.740 \\
                            & STGNN & 0.0041 & 0.0641 & 0.903 \\
\bottomrule
\end{tabular}
\end{table}

\begin{itemize}
    \item \textbf{GH2 (2020)}:
The vanilla RNN attains near-perfect accuracy ($R^{2}=0.958$) on the relatively simple six-variable setting, comfortably outperforming the directed STGNN.  Here the microclimate is governed mainly by smooth external forcing; explicit directional edges add little beyond the temporal cues already captured by the RNN.

\item \textbf{GH4 (2024)}: Adding PAR and \mbox{CO\textsubscript{2}} sensors increases both dimensionality and interaction complexity; the STGNN now surpasses the RNN by about 16 percentage points in $R^{2}$ (0.905 vs 0.740) while halving the MSE.  Attention heads exploit directed links such as \textit{OUT\_PAR\,$\rightarrow$\,G4\_PAR} and \textit{G4\_PAR\,$\rightarrow$\,G4\_Temp}, which a purely temporal model cannot encode explicitly.
\end{itemize}
 
GH2’s variables form a nearly linear control chain (outside $\rightarrow$ inside) with limited cross-talk.  In contrast, GH4 exhibits multiple feedback loops: light drives photosynthesis, lowering internal \mbox{CO\textsubscript{2}}, which in turn affects transpiration and humidity, all interacting with ventilation events.  The graph attention mechanism can re-weight these concurrent pathways on the fly, whereas the RNN must learn an entangled representation from sequences alone, hence its lower accuracy.

For completeness, Table~\ref{tab:gh2_seasonal} in Appendix~\ref{secAppendixSeasonal} reports GH2 Summer and Winter scores.  The RNN retains a narrow edge in both cases, confirming that when the variable set is modest and control actions are well understood, a lightweight temporal network suffices.

\begin{figure}[H]
\centering
\caption{GH2 Autumn. Observed vs.\ predicted temperature.}
\label{fig:gh2_autumn_scatter}
\begin{subfigure}{0.49\textwidth}
    \centering
    \includegraphics[width=\linewidth]{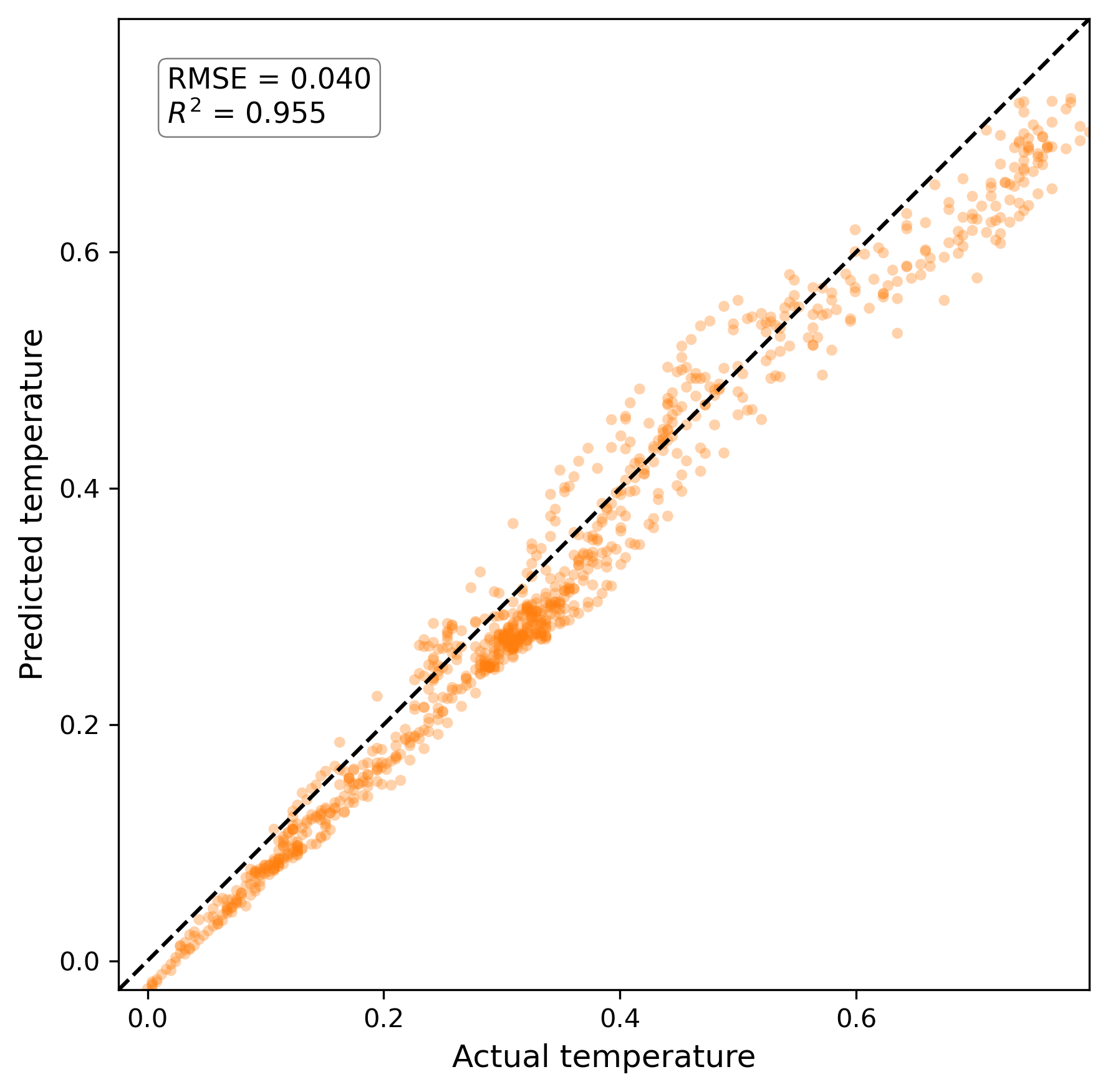}
    \caption{RNN scatter}
\end{subfigure}
\hfill
\begin{subfigure}{0.49\textwidth}
    \centering
    \includegraphics[width=\linewidth]{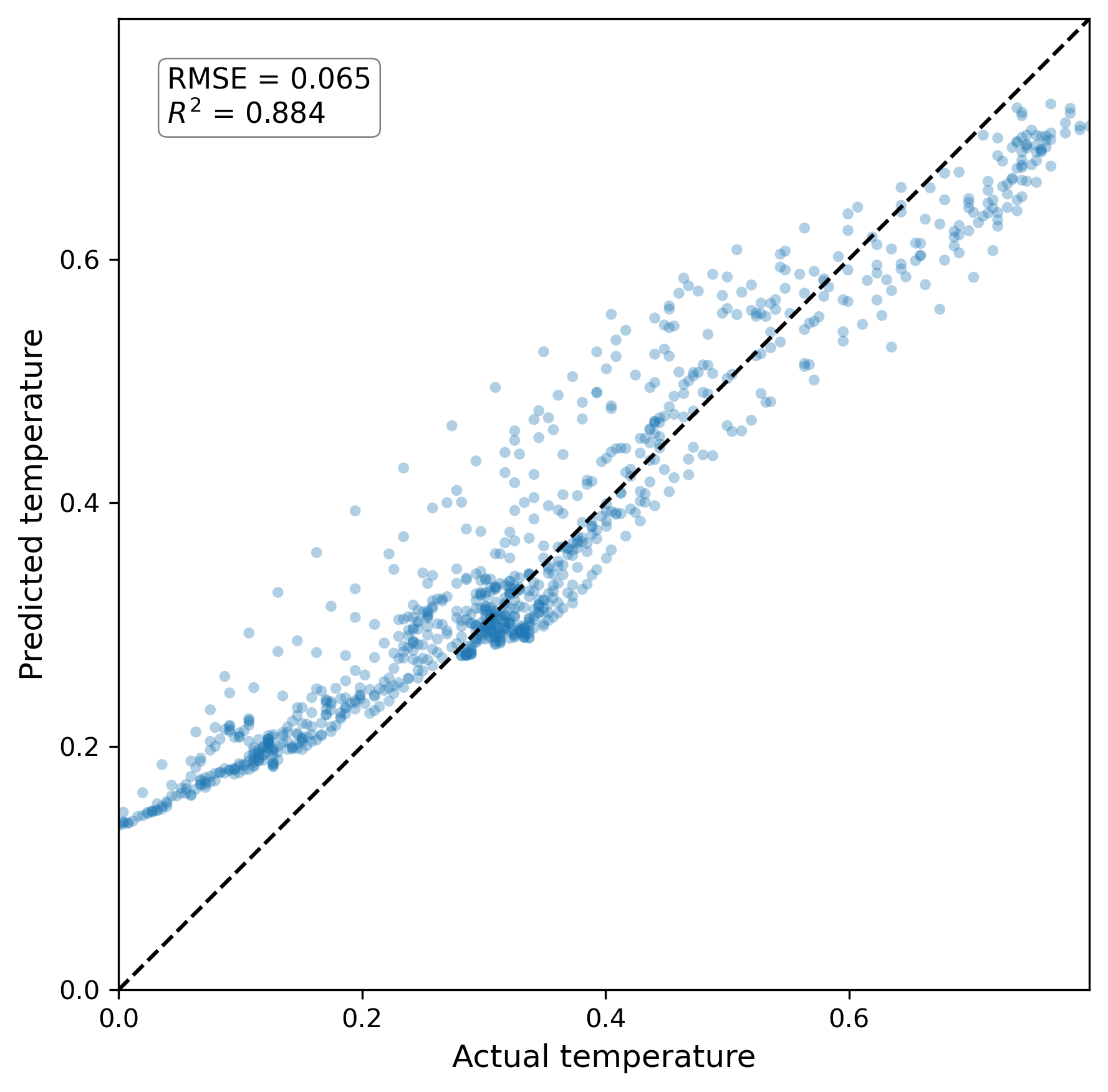}
    \caption{STGNN scatter}
\end{subfigure}
\end{figure}

\begin{figure}[H]
  \centering
  \caption{GH4~Autumn. Observed vs.\ predicted temperature.}
  \begin{subfigure}[b]{0.49\textwidth}
    \centering
    \includegraphics[width=\linewidth]{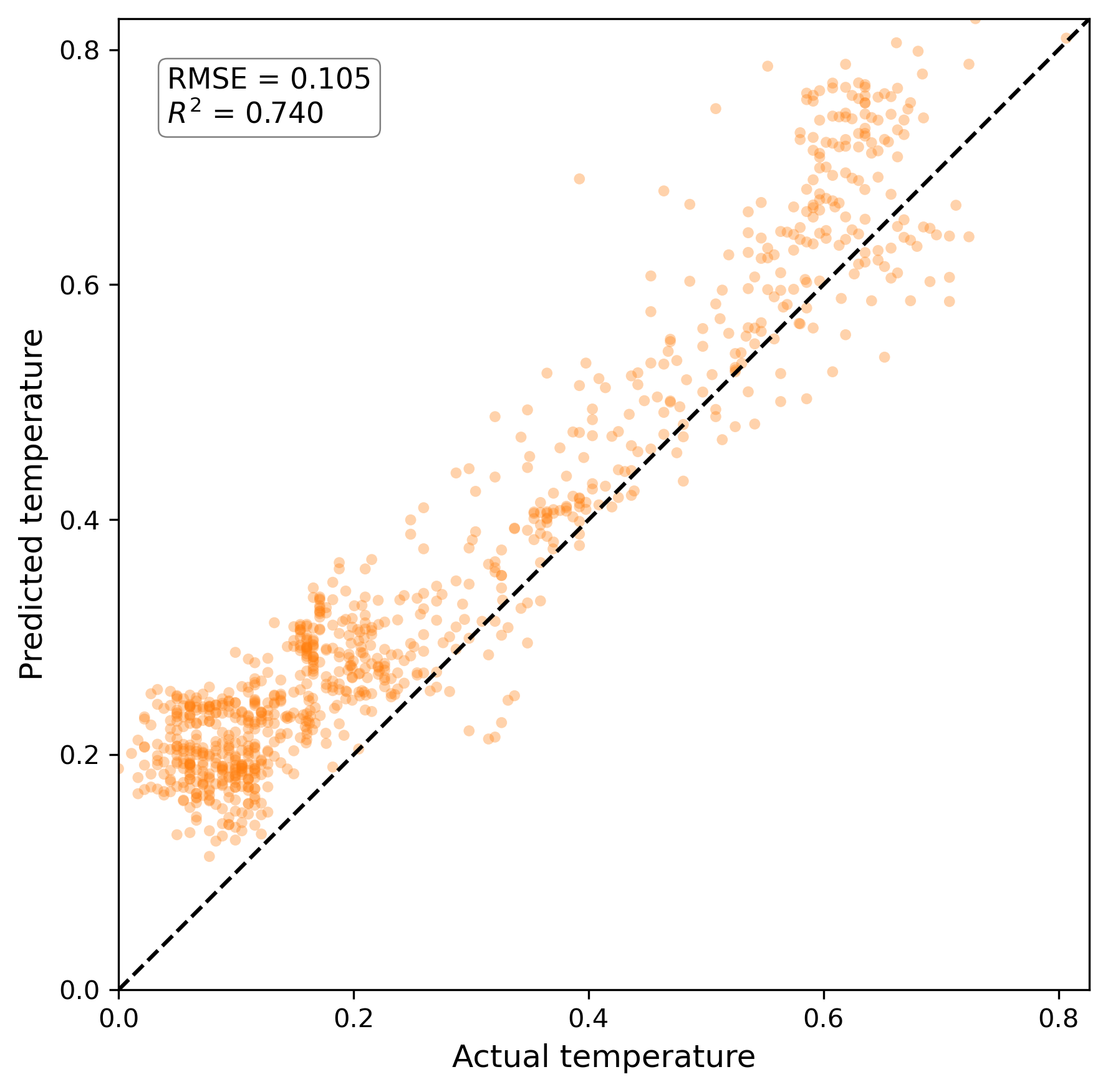}
    \caption{RNN}
  \end{subfigure}
  \hfill  % small vertical gap
  \begin{subfigure}[b]{0.49\textwidth}
    \centering
    \includegraphics[width=\linewidth]{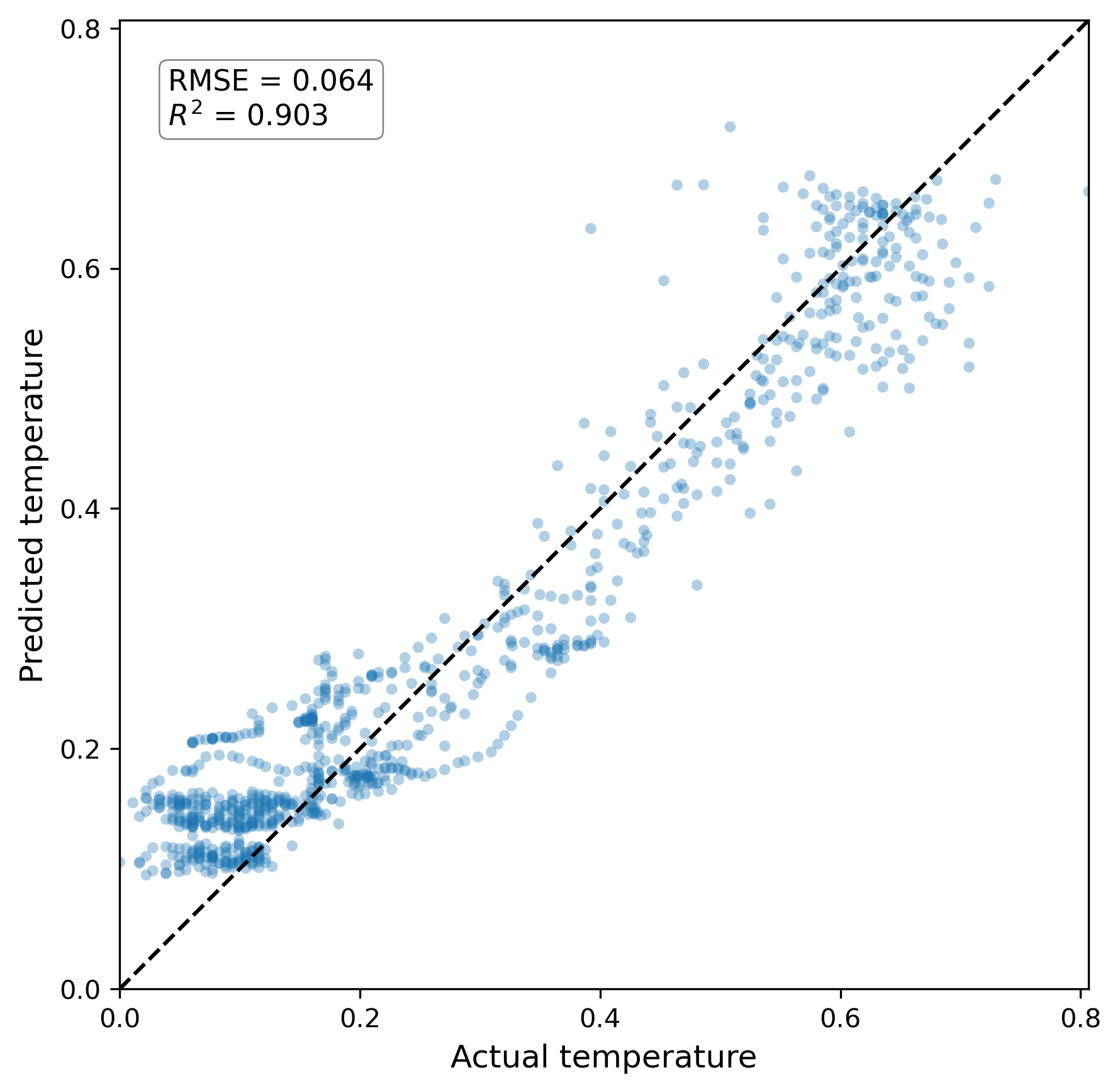}
    \caption{STGNN}
  \end{subfigure}
  \label{fig:gh4_autumn_scatter}
\end{figure}

\begin{figure}[H]
  \centering
  \caption{GH2~Autumn. Time series of observed and predicted temperature.}
  \begin{subfigure}[b]{0.95\textwidth}
    \centering
    \includegraphics[width=\linewidth]{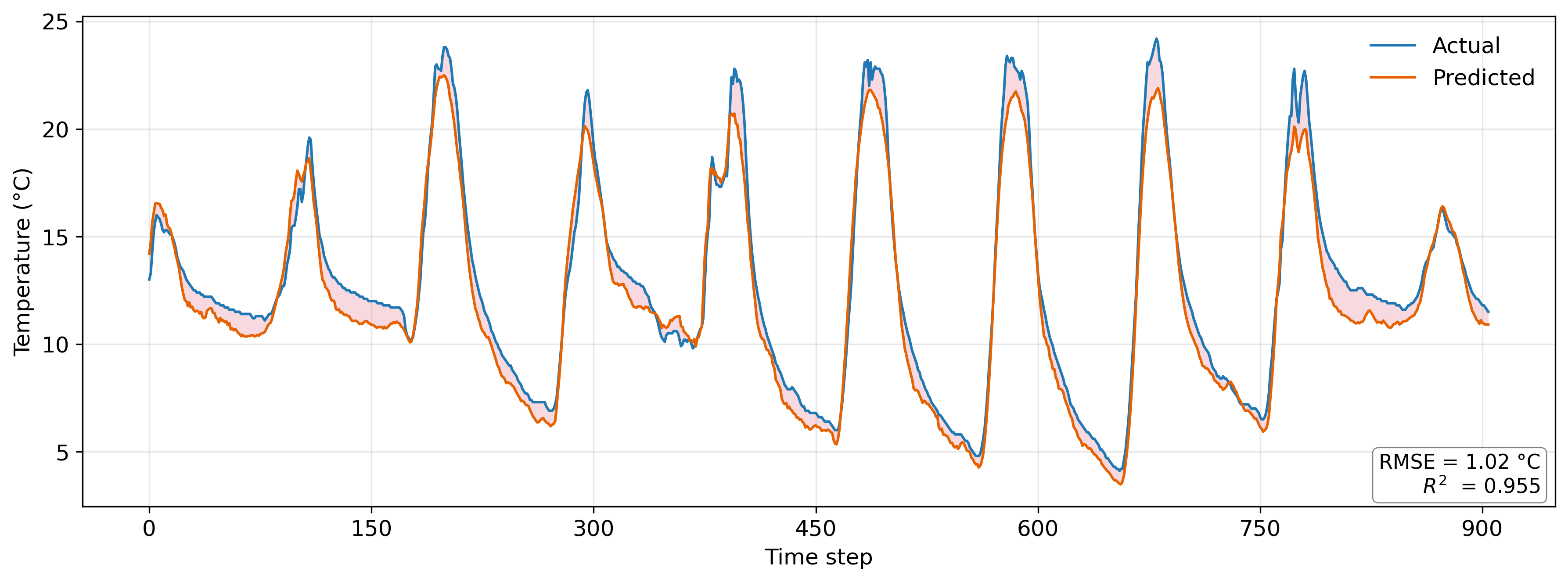}
    \caption{RNN}
  \end{subfigure}
  \vspace{6pt}  % small vertical gap
  \begin{subfigure}[b]{0.95\textwidth}
    \centering
    \includegraphics[width=\linewidth]{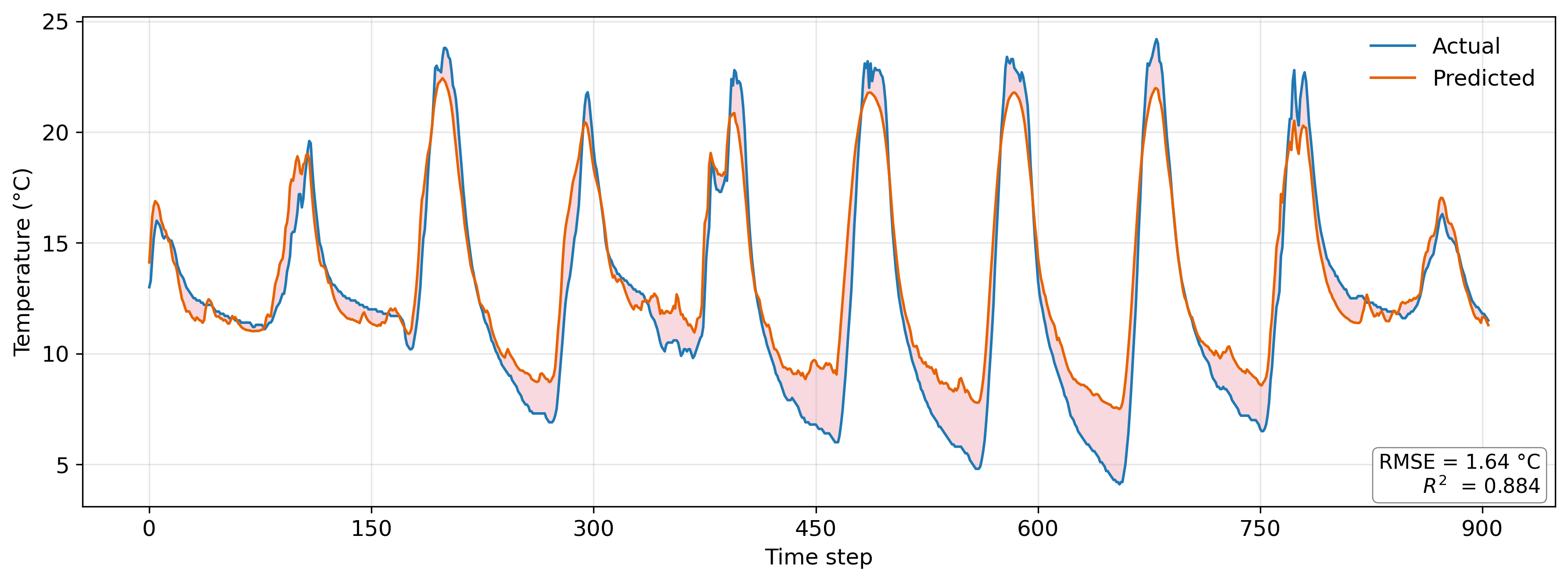}
    \caption{STGNN}
  \end{subfigure}
  \label{fig:gh2_autumn_time}
\end{figure}

\begin{figure}[H]
  \centering
  \caption{GH4~Autumn. Time series of observed and predicted temperature.}
  \begin{subfigure}[b]{0.95\textwidth}
    \centering
    \includegraphics[width=\linewidth]{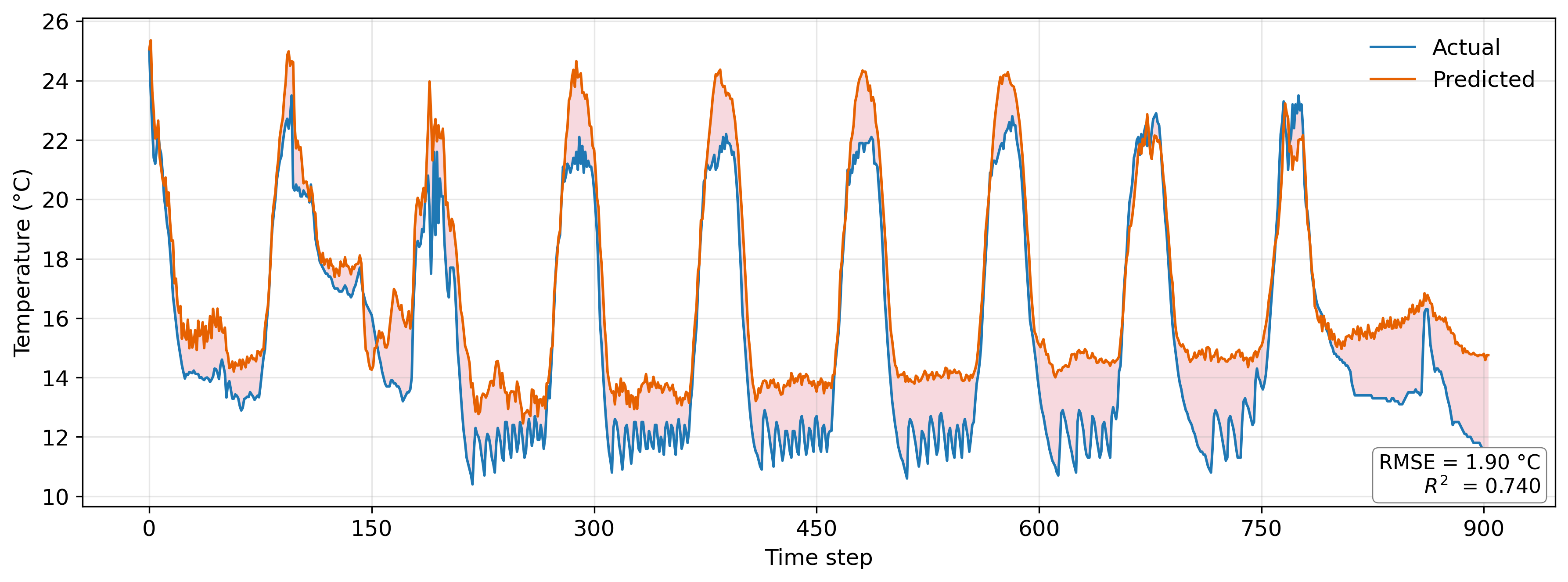}
    \caption{RNN}
  \end{subfigure}
  \vspace{6pt}  % small vertical gap
  \begin{subfigure}[b]{0.95\textwidth}
    \centering
    \includegraphics[width=\linewidth]{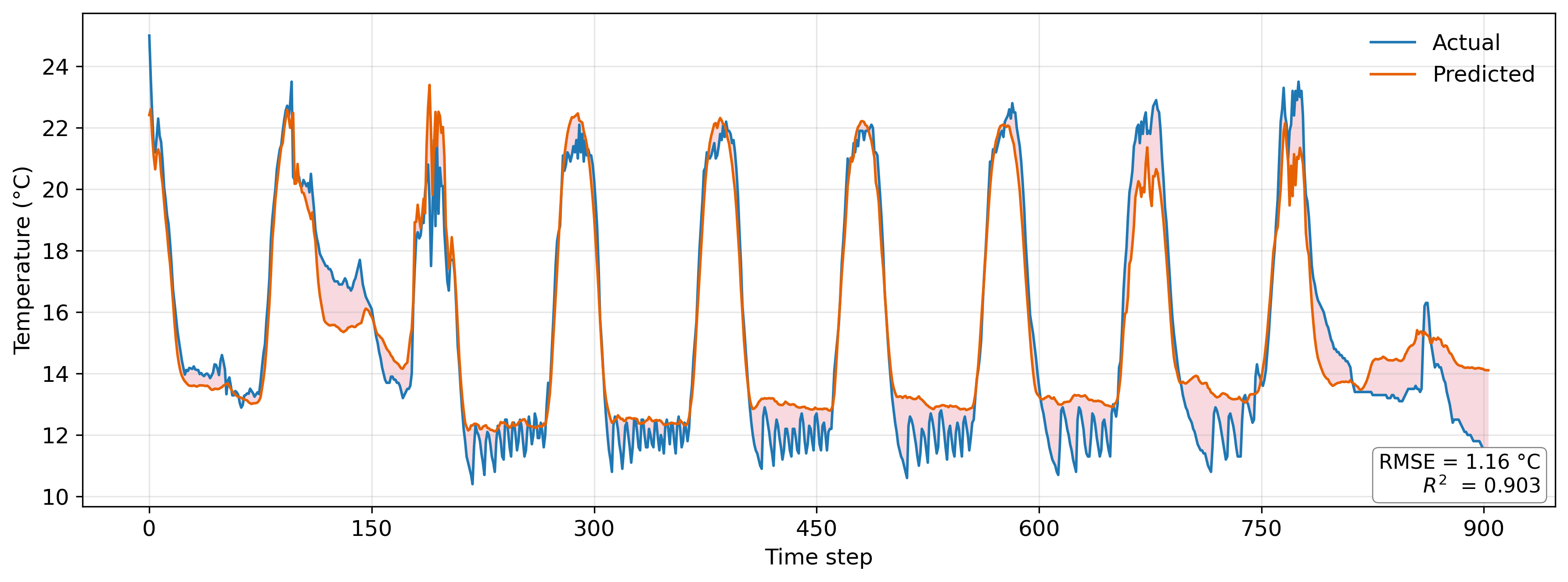}
    \caption{STGNN}
  \end{subfigure}
  \label{fig:gh4_autumn_time}
\end{figure}

Figures~\ref{fig:gh4_autumn_scatter} and\ref{fig:gh4_autumn_time} visualise the GH4 test results, while Figures~\ref{fig:gh2_autumn_scatter} and \ref{fig:gh2_autumn_time} provide the corresponding GH2 Autumn comparison. On the time series plots, inverse scaling have been performed to have Celsius degrees on the y-axis.

Figure~\ref{fig:gh4_autumn_scatter} contrasts the two 1:1 scatter plots for GH4.  The STGNN cloud is tightly clustered along the diagonal, with only modest spread at the lowest temperatures, whereas the RNN points disperse into a pronounced “fan”: the network tends to over-predict cool nights and under-predict the hottest daytime spikes.  The stacked traces in Fig.~\ref{fig:gh4_autumn_time} confirm this pattern, STGNN rises and falls almost in phase with measurements, capturing sharp midday peaks and steep nocturnal cooldowns, whereas the RNN lags and drifts, explaining its higher RMSE and lower $R^{2}$.

In contrast, GH2 Autumn plots (Figures~\ref{fig:gh2_autumn_scatter} and \ref{fig:gh2_autumn_time}) show both models hugging the diagonal and tracking the time series closely; the RNN’s slight advantage is visible only in a tighter scatter and marginally smaller peak-to-peak residuals.  The absence of PAR/CO\textsubscript{2} feedback loops leaves little room for the graph attention mechanism to improve upon the temporal baseline.

Overall, these results substantiate the central claim: \emph{directional graphs pay off once sensor richness amplifies interaction pathways}, while a simpler RNN already saturates performance in low-dimensional, weakly coupled settings.

\section{Discussion}
\label{sec: Discussion}

The autumn experiments on two greenhouses of increasing dimensionality illuminate when a purely temporal network is sufficient and when an explicitly directed graph becomes more useful. In the six-variable GH2 data set, the vanilla RNN already explains more than 95\% of the temperature variance and leaves little residual structure for a graph network to exploit. The microclimate can be approximated as a nearly one-way control chain in which external temperature and radiation drive internal conditions with limited feedback, so the chronological context contained in 96 time steps is enough for the RNN to learn the underlying dynamics. 
By contrast, the eight-variable PV house (GH4) adds PAR and CO\textsubscript{2} sensors that introduce two new, strongly coupled, feedback loops: (i) light raises PAR inside the house, boosting photosynthesis and thus lowering internal CO\textsubscript{2}; (ii) ventilation events triggered by heat or humidity instantaneously mix outside and inside air, altering both CO\textsubscript{2} and RH. These loops generate temperature excursions that depend not only on the magnitude of each driver but on their \emph{direction} and relative timing. The GAT layer can re-weight the edges \textit{OUT\_PAR,$\rightarrow$,G4\_PAR}, \textit{G4\_PAR,$\rightarrow$,G4\_Temp} and \textit{G4\_CO2,$\leftrightarrow$,G4\_Temp} on the fly, whereas the RNN is forced to embed all interactions into a single recurrent state. The result is a significant gain in $R^{2}$ and drop in MSE for the STGNN, confirming that \emph{directional graphs pay off once sensor richness amplifies interaction pathways}.

Several practical lessons follow. First, model choice should be driven by anticipated variable coupling rather than by the generic appeal of “spatial” deep learning: if only a handful of weakly interacting drivers are measured, a lightweight RNN is likely to saturate performance. Second, the incremental cost of a graph network is modest, one additional GAT layer with $K=4$ heads adds fewer than 30 k parameters, so upgrading to an STGNN is advisable whenever new sensors or control rules create non-trivial feedback. Third, graph quality matters: the manually defined edges used here relied on fundamental thermodynamic reasoning and produced stable training; future deployments should test data-driven edge discovery or adaptive pruning to avoid spurious links when variable sets grow further.

Limitations remain. The directed graphs capture variable-to-variable causality but not spatial heterogeneity within the greenhouse (e.g.\ temperature gradients along the crop rows) nor operational events such as window openings or heating pulses, which would require either additional sensors or hybrid physical–empirical coupling. Moreover, even the richer GH4 data span only six weeks; longer records across seasons are needed to quantify generalisation.

\section{Conclusions and future developments}
\label{sec: conclusions}

This study compared Recurrent Neural Networks (RNNs) and directed Spatio-Temporal Graph Neural Networks (STGNNs) for quarter-hour greenhouse, temperature prediction. Two data sets of growing dimensionality were analysed: a six-variable 2020 installation (GH2) and an eight-variable 2024 PV house (GH4).

When drivers form an almost one-way chain, as in GH2, a vanilla RNN explains more than 95\% of the variance and the extra graph structure yields no gain. In GH4, however, the addition of PAR and CO\textsubscript{2} sensors creates coupled feedback loops that a purely temporal model cannot disentangle. Introducing one GAT layer with $K=4$ heads raises the STGNN to $R^{2}=0.905$ and cuts the MSE by half, while adding only approximately 30k parameters, negligible overhead by modern deep-learning standards. \emph{Thus, graph-based models become worthwhile once richer sensing introduces directional cross-talk, whereas a plain RNN suffices for low-dimensional, weakly coupled settings.}
The practical takeaway is clear: model choice should reflect expected coupling complexity, not architectural fashion. Installations with a few loosely interacting variables need nothing more elaborate than an RNN.
Building on the insights gained from a single geographic region, future work will actively extend this analysis across diverse climates, greenhouse designs, and sensor setups to robustly demonstrate the broader applicability and scalability of the STGNN approach.

This work is an early step toward the REGACE project’s digital-twin vision for PV-integrated greenhouses. Although we focus here on temperature, the graph framework readily accommodates new nodes. Future work will embed PV-panel characteristics, alternative panel layouts, and crop-growth variables, assessing their joint impact on energy yield and microclimate. Coupling these empirical models with detailed physical simulations will enable rapid design screening and, ultimately, integrated control of agrivoltaic greenhouses, supporting wider adoption while ensuring optimal conditions for both power generation and crop production.

\section*{CRediT authorship contribution statement}
\textbf{Emiliano Seri:} Writing – original draft \& editing, Visualization, Validation, Software, Methodology, Investigation, Formal analysis, Data curation, Conceptualization. \textbf{Marcello Petitta:} Writing – original draft \& editing, Conceptualization, Investigation, Formal analysis, Methodology, Resources, Funding acquisition, Supervision. 
\textbf{Chryssoula Papaioannou:} Data curation. 
\textbf{Nikolaos Katsoulas:} Data curation. 
\textbf{Cristina Cornaro:} Writing – original draft \& editing, Conceptualization, Formal analysis, Funding acquisition, Supervision.

\section*{Data availability}

The authors do not have permission to share data.

\section*{Declaration of competing interest}
The authors declare that they have no known competing financial interests or personal relationships that could have appeared to influence the work reported in this paper.

\section*{Acknowledgments}

This research was carried out in the framework of the REGACE project funded by the European Union under Grant Agreement No 101096056. Views and opinions expressed are however those of the author(s) only and do not necessarily reflect those of the European Union or CINEA. Neither the European Union nor the granting authority can be held responsible for them.

\bibliography{MAIN}

\appendix
\section{GH2 Summer and Winter results}
\label{secAppendixSeasonal}
%------------------------------------------------------------------

\subsection*{A.1  Test metrics}

\begin{table}[H]
\centering
\caption{GH2 Summer and Winter. Test-set performance ($T\!=\!96$, 32\,epochs, batch 96)}
\label{tab:gh2_seasonal}
\begin{tabular}{lcccc}
\toprule
Season & Model   & MSE   & RMSE  & $\bm{R^2}$ \\
\midrule
\multirow{2}{*}{Winter} & RNN   & 0.0019 & 0.0433 & 0.957 \\
                        & STGNN & 0.0019 & 0.0441 & 0.955 \\[2pt]
\multirow{2}{*}{Summer} & RNN   & 0.0061 & 0.0781 & 0.883 \\
                        & STGNN & 0.0062 & 0.0785 & 0.882 \\
\bottomrule
\end{tabular}
\end{table}

%------------------------------------------------------------------
\subsection*{A.2  Winter figures}

\begin{figure}[H]
    \centering
    \caption{GH2 temperatures inside and outside the greenhouse for the winter period}
\label{fig: wintertemp}
\includegraphics[width=1\textwidth]{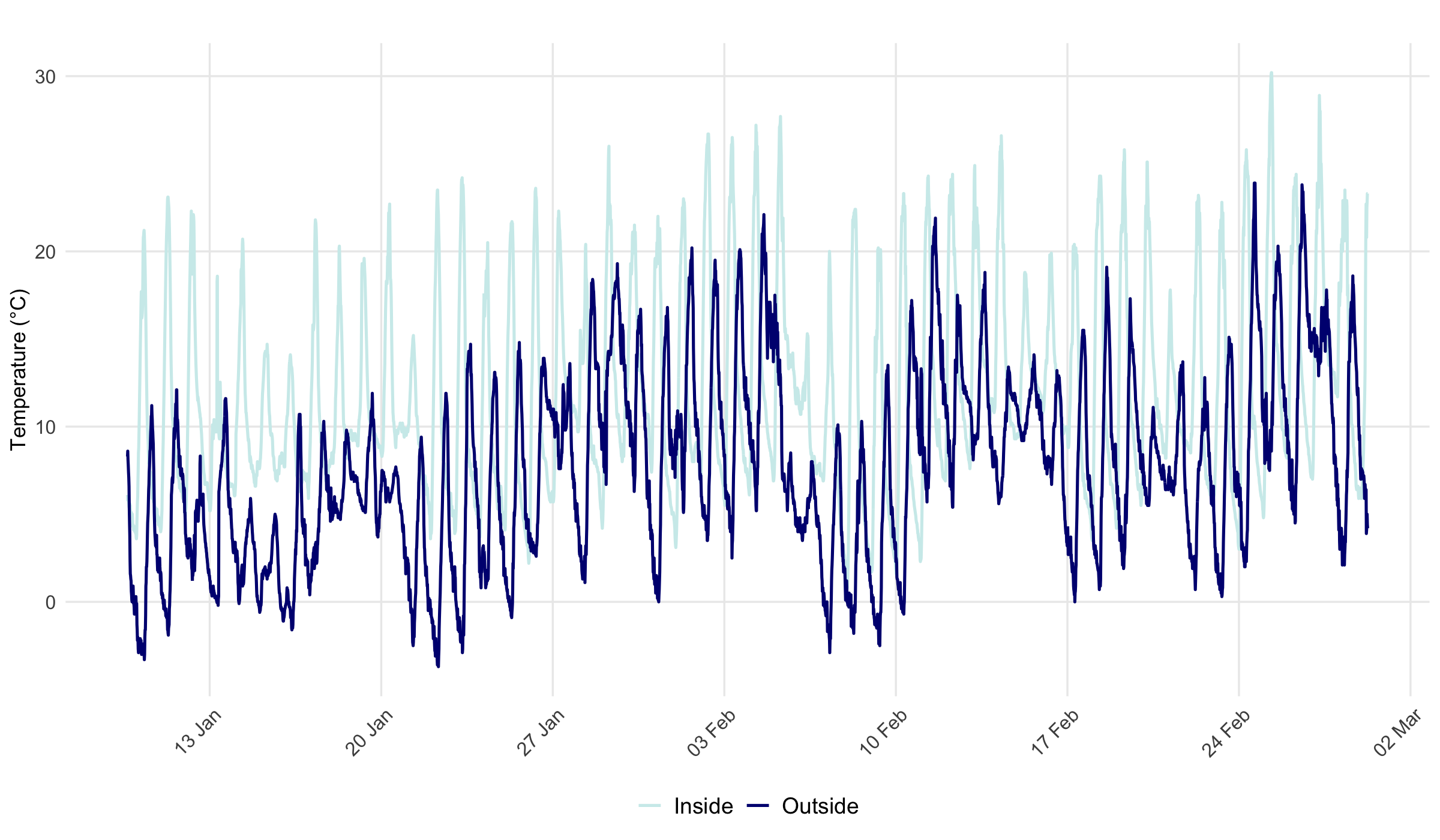}
\end{figure}

\begin{figure}[H]
\centering
\caption{GH2 Winter. Observed vs.\ predicted temperature.}
\label{fig:gh2_winter_scatter}
\begin{subfigure}{0.49\textwidth}
    \centering
    \includegraphics[width=\linewidth]{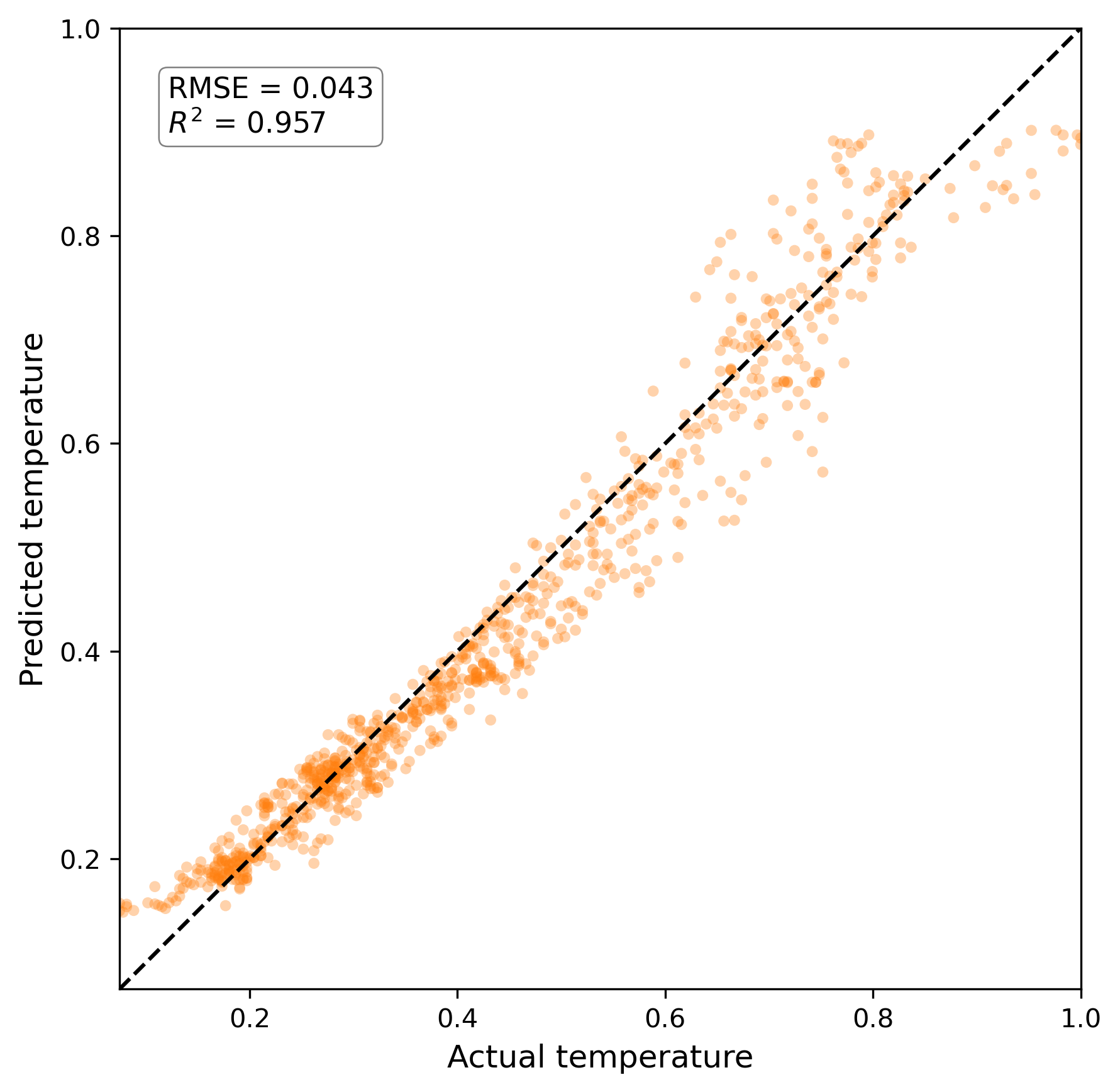}
    \caption{RNN scatter}
\end{subfigure}
\hfill
\begin{subfigure}{0.49\textwidth}
    \centering
    \includegraphics[width=\linewidth]{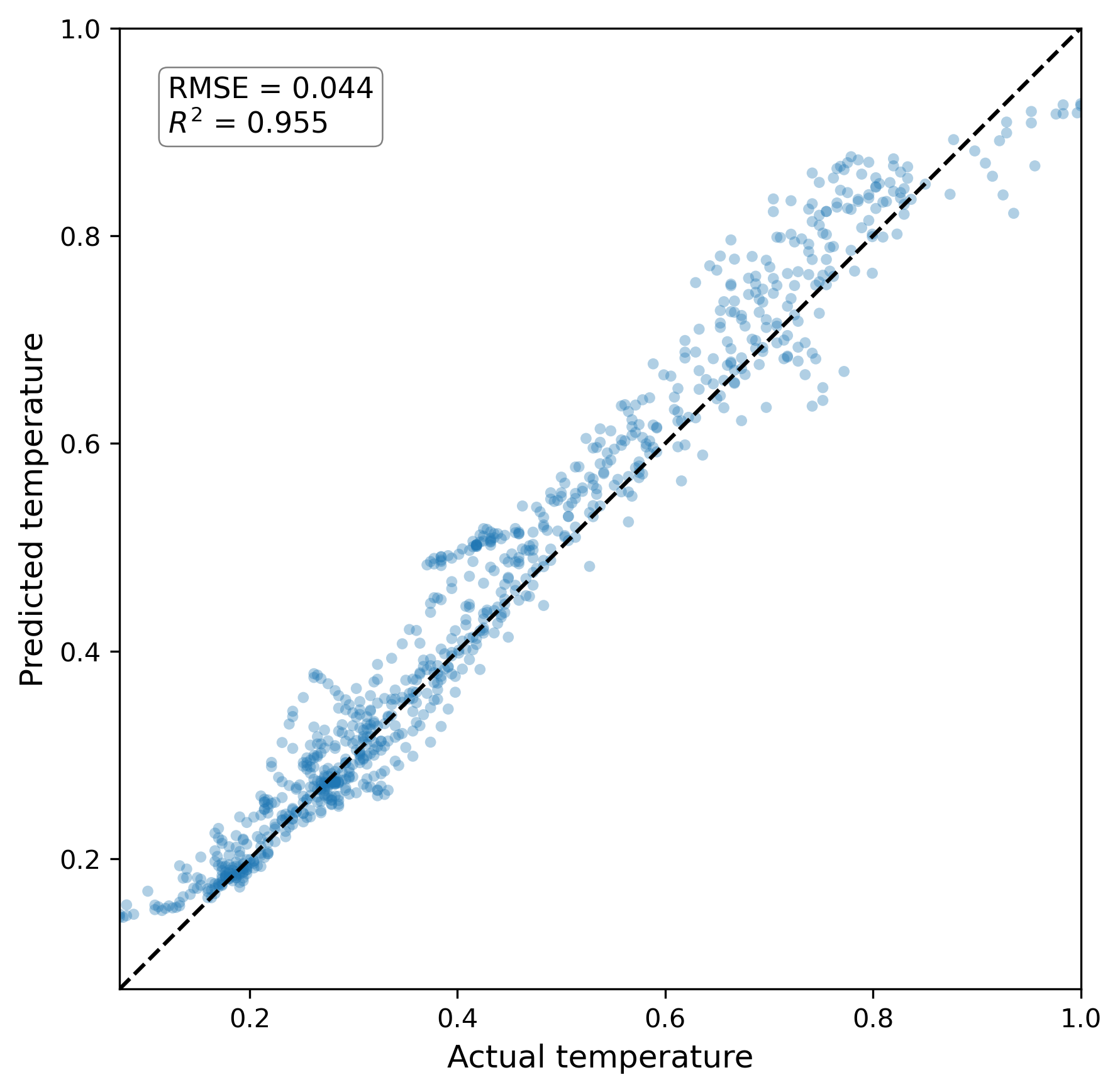}
    \caption{STGNN scatter}
\end{subfigure}
\end{figure}

%------------------------------------------------------------------
\subsection*{A.3  Summer figures}

\begin{figure}[H]
    \centering
    \caption{GH2 Temperatures inside and outside the greenhouse for the summer period}
\label{fig: summertemp}
\includegraphics[width=1\textwidth]{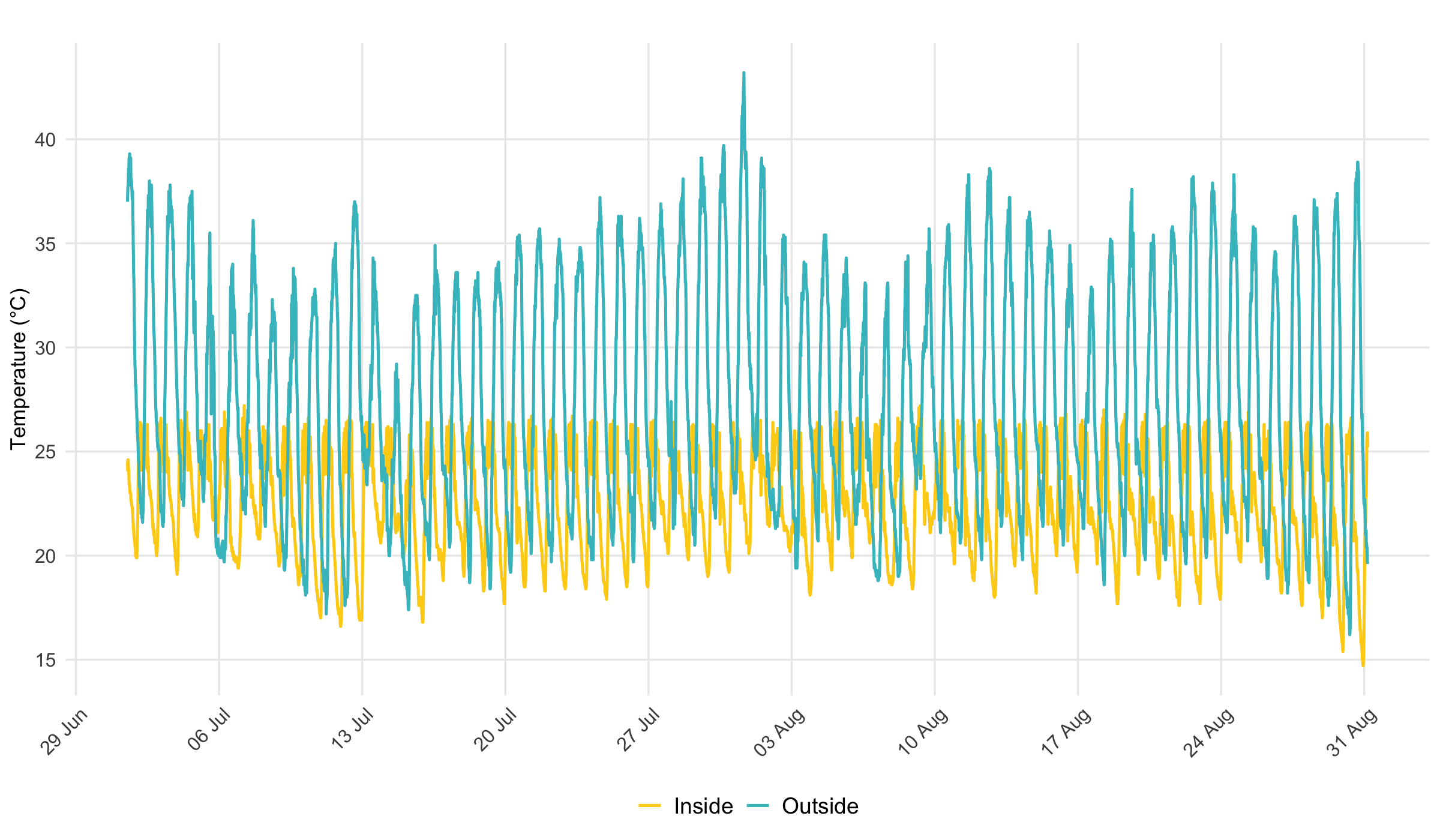}
\end{figure}

\begin{figure}[H]
\centering
\caption{GH2 Summer. Observed vs.\ predicted temperature.}
\label{fig:gh2_summer_scatter}
\begin{subfigure}{0.49\textwidth}
    \centering
    \includegraphics[width=\linewidth]{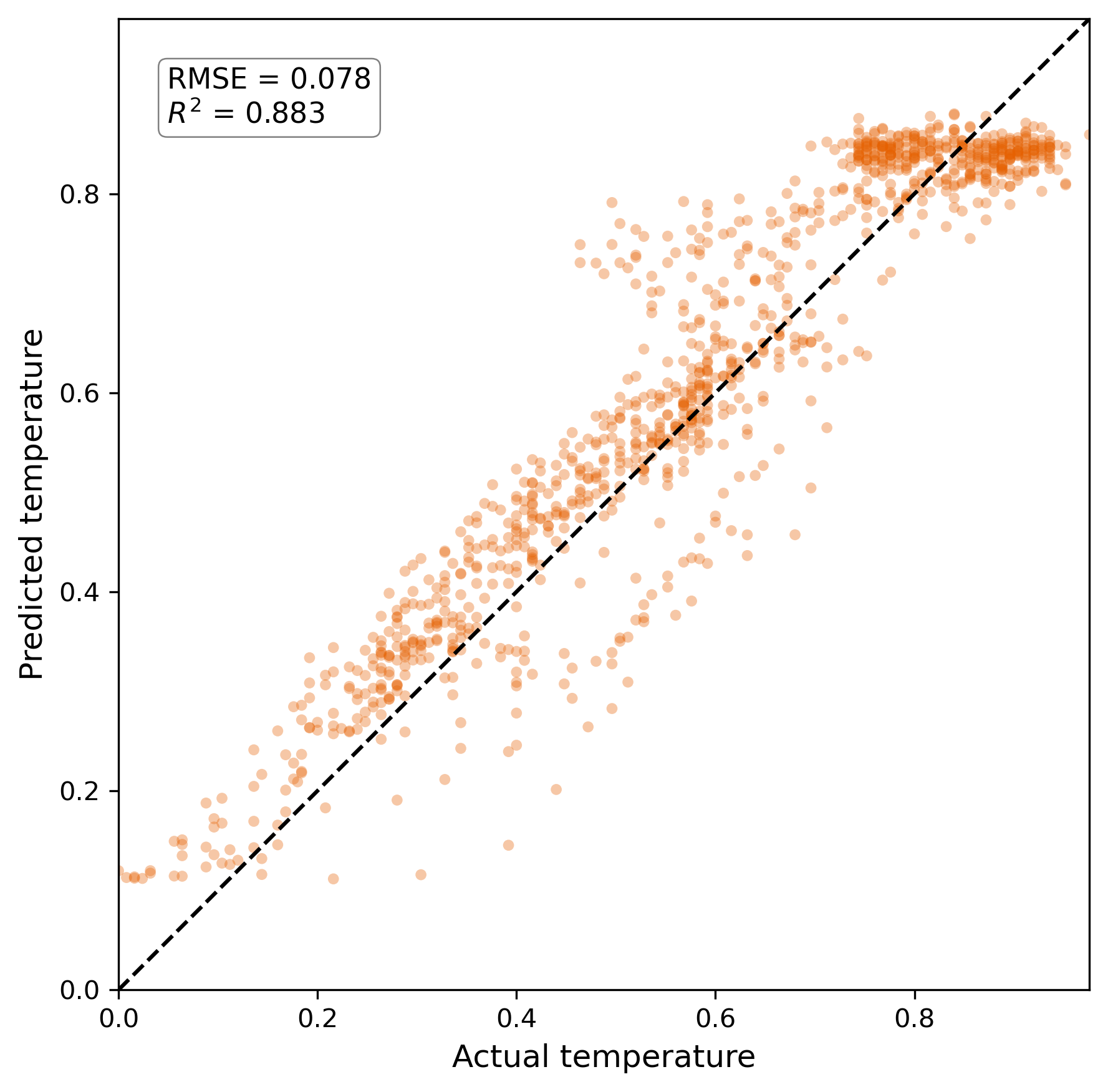}
    \caption{RNN scatter}
\end{subfigure}
\hfill
\begin{subfigure}{0.49\textwidth}
    \centering
    \includegraphics[width=\linewidth]{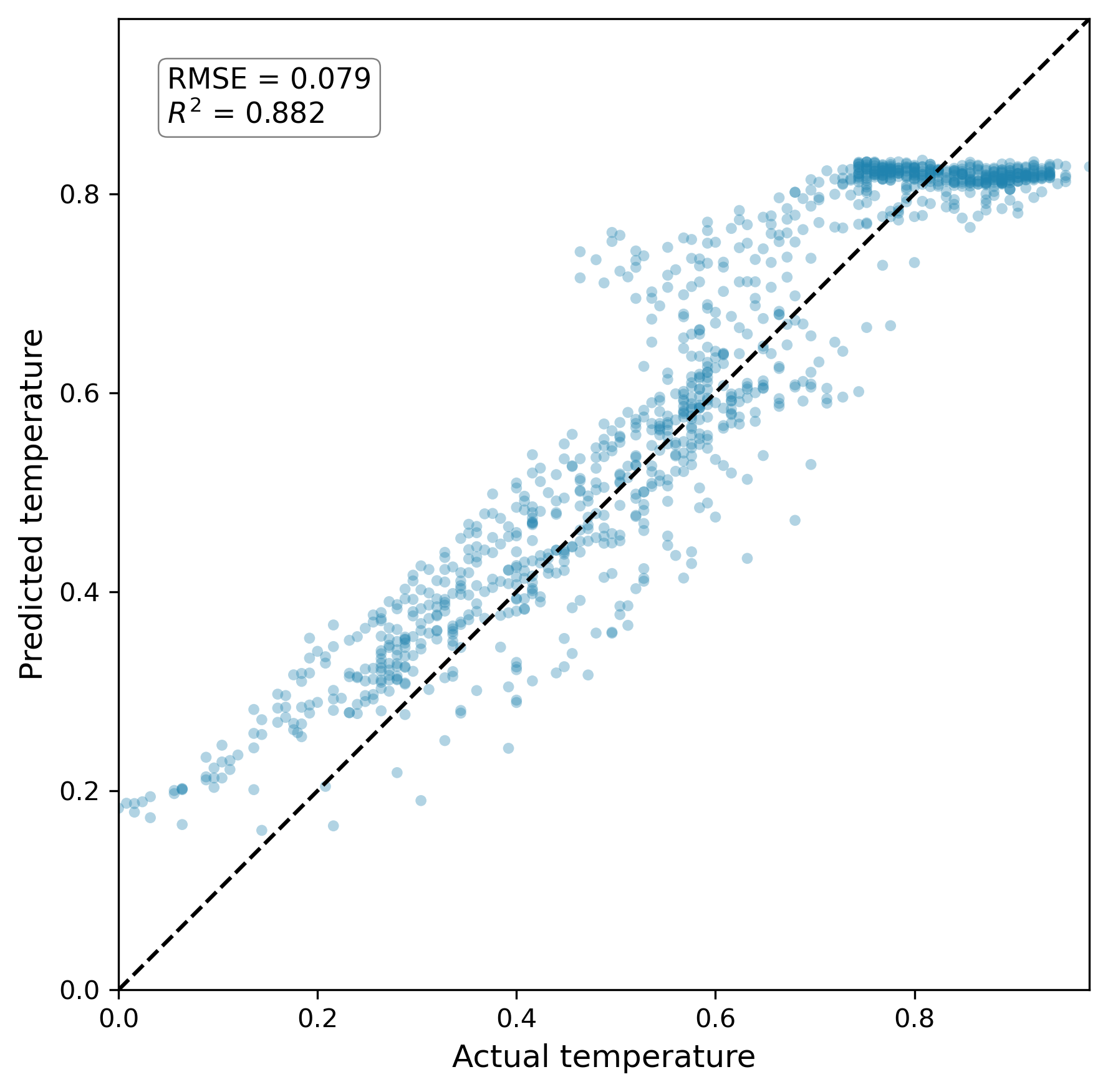}
    \caption{STGNN scatter}
\end{subfigure}
\end{figure}

\end{document}